\documentclass{article}

\usepackage[preprint]{corl_2023} % Uncomment for pre-prints (e.g., arxiv); This is like ``final'', but will remove the CORL footnote.

\usepackage{subcaption}
\usepackage{graphicx}
\usepackage{amsmath}
\usepackage{amssymb}
\usepackage{booktabs}
\usepackage{multirow}

\usepackage{enumitem}
\usepackage{listings}
\usepackage[T1]{fontenc}
\usepackage[scaled=.8]{beramono}   % Scale texttt. Default is .9

\usepackage{soul}
\usepackage{xspace}
\sethlcolor{white}
\usepackage{amsmath}

\DeclareMathOperator*{\argmin}{arg\,min}

\newcommand{\MODELNAME}[0]{{\fontfamily{txtt}\selectfont {LCTGen}}\xspace}
\newcommand{\GENERATORNORMAL}[0]{{\fontfamily{txtt}\selectfont\textcolor{black} {Generator}}\xspace}
\newcommand{\GENERATOR}[0]{{\small\fontfamily{txtt}\selectfont\textcolor{black} {Generator}}\xspace}
\newcommand{\ENCODERNORMAL}[0]{{\fontfamily{txtt}\selectfont\textcolor{black} {Encoder}}\xspace}
\newcommand{\ENCODER}[0]{{\small\fontfamily{txtt}\selectfont\textcolor{black} {Encoder}}\xspace}
\newcommand{\INTERPRETERNORMAL}[0]{{\fontfamily{txtt}\selectfont\textcolor{black} {Interpreter}}\xspace}
\newcommand{\INTERPRETER}[0]{{\small\fontfamily{txtt}\selectfont\textcolor{black} {Interpreter}}\xspace}
\newcommand{\RETRIEVALNORMAL}[0]{{\fontfamily{txtt}\selectfont\textcolor{black} {Retrieval}}\xspace}
\newcommand{\RETRIEVAL}[0]{{\small\fontfamily{txtt}\selectfont\textcolor{black} {Retrieval}}\xspace}
\newcommand{\RNum}[1]{\uppercase\expandafter{\romannumeral #1\relax}}

\lstdefinestyle{prompt}{
    language=sh,
    basicstyle=\footnotesize\ttfamily\color{white},
    keywordstyle=\color{white},
    emphstyle=\color{black},
    commentstyle=\color{green},
    showstringspaces=false,
    numbers=left,
    numberstyle=\footnotesize\color{white},
    frame=single,
    framesep=5pt,
    rulecolor=\color{black},
    xleftmargin=15pt,
    framexleftmargin=10pt,
    backgroundcolor=\color{black},
    moredelim=[il][\textcolor{black}]{$ $},
    moredelim=[is][\textcolor{black}]{\%\%}{\%\%},
}

\usepackage[capitalize]{cleveref}
\crefname{section}{Sec.}{Secs.}
\Crefname{section}{Section}{Sections}
\Crefname{table}{Table}{Tables}
\crefname{table}{Tab.}{Tabs.}

\renewcommand{\baselinestretch}{0.9}

\title{Language Conditioned Traffic Generation}

\author{
  \textbf{Shuhan Tan}$^1$ \quad \textbf{Boris Ivanovic}$^2$ \quad \textbf{Xinshuo Weng}$^2$ \\
  \textbf{Marco Pavone}$^2$ \quad \textbf{Philipp Kr\"ahenb\"uhl}$^1$ \\
  $^1$UT Austin \quad $^2$NVIDIA\\
}

\begin{document}

\maketitle
%===============================================================================
\vspace{-0.5cm}
\begin{abstract}
Simulation forms the backbone of modern self-driving development.
Simulators help develop, test, and improve driving systems without putting humans, vehicles, or their environment at risk.
However, simulators face a major challenge: They rely on realistic, scalable, yet interesting content.
While recent advances in rendering and scene reconstruction make great strides in creating static scene assets, modeling their layout, dynamics, and behaviors remains challenging.
In this work, we turn to language as a source of supervision for dynamic traffic scene generation.
Our model, \MODELNAME, combines a large language model with a transformer-based decoder architecture that selects likely map locations from a dataset of maps, produces an initial traffic distribution, as well as the dynamics of each vehicle.
\MODELNAME outperforms prior work in both unconditional and conditional traffic scene generation in-terms of realism and fidelity.
Code and video will be available at \url{https://ariostgx.github.io/lctgen}.
\end{abstract}

\section{Introduction}
\label{sec:intro}
Driving simulators stand as a cornerstone in self-driving development.
They aim to offer a controlled environment to mimic real-world conditions and produce critical scenarios at scale.
Towards this end, they need to be highly \textit{realistic} (to capture the complexity of real-world environments), \textit{scalable} (to produce a diverse range of scenarios without excessive manual effort), and able to create \textit{interesting} traffic scenarios (to test self-driving agents under different situations).

In this paper, we turn to natural language as a solution.
Natural language allows practitioners to easily articulate interesting and complex traffic scenarios through high-level descriptions.
Instead of meticulously crafting the details of each individual scenario, language allows for a seamless conversion of semantic ideas into simulation scenarios at scale.
To harness the capacity of natural language, we propose \MODELNAME. 
\MODELNAME takes as input a natural language description of a traffic scenario, and outputs traffic actors' initial states and motions on a compatible map. 
As we will show in Section~\ref{sec:exp}, \MODELNAME generates realistic traffic scenarios that closely adhere to a diverse range of natural language descriptions, including detailed crash reports~\cite{national2016crash}.

\begin{figure*}[!t]
\centering
% \vspace{-5em}
\includegraphics[width=0.9\linewidth]{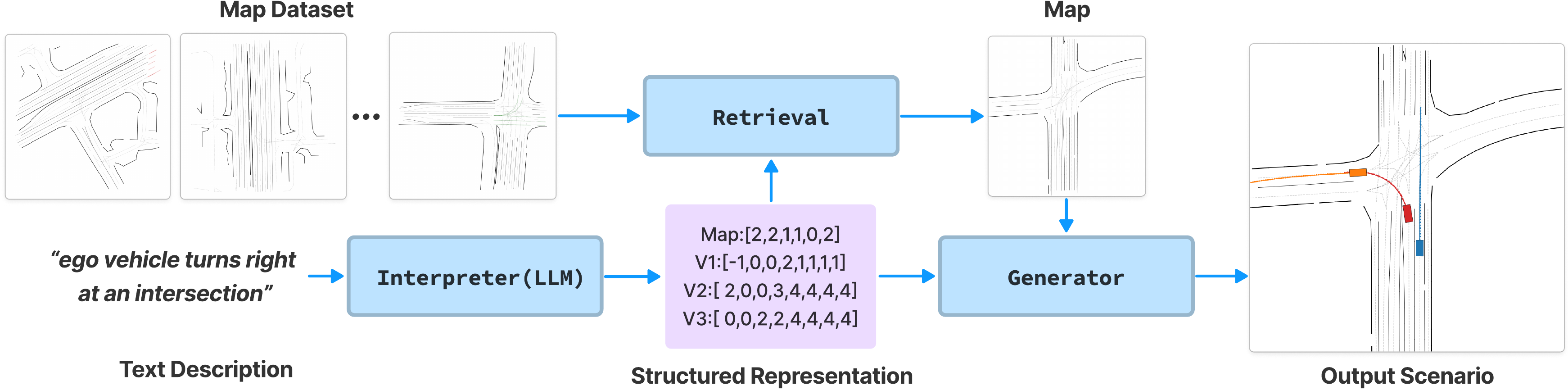}
% \vspace{-3em}
\caption{Overview of our \MODELNAME model.}
\vspace{-1.5em}
\label{fig:pipeline}
\end{figure*}

The major challenge of language-conditioned traffic generation is the absence of a shared representation between language and traffic scenarios. 
Furthermore, there are no paired language-traffic datasets to support learning such a representation.
To address these challenges, \MODELNAME (see Figure~\ref{fig:pipeline}) uses a scenario-only dataset and a Large Language Model (LLM).
\MODELNAME has three modules: \INTERPRETER, \GENERATOR and \ENCODER.
Given any user-specified natural language query, the LLM-powered \INTERPRETER converts the query into a compact, structured representation.
\INTERPRETER also retrieves an appropriate map that matches the described scenario from a real-world map library.
Then, the \GENERATOR takes the structured representation and map to generate realistic traffic scenarios that accurately follow the user's specifications.
Also, we design the \GENERATOR as a query-based Transformer model~\cite{NIPS2017_3f5ee243}, which efficiently generates the full traffic scenario in a single pass.

This paper presents three main contributions:
\begin{enumerate}[leftmargin=0.8cm]
\vspace{-0.2cm}
    \itemsep-0.2em 
    \item We introduce \MODELNAME, a first-of-its-kind model for language-conditional traffic generation. 
    \item We devise a method to harness LLMs to tackle the absence of language-scene paired data.
    \item \MODELNAME exhibits superior realism and controllability over prior work. We also show \MODELNAME can be applied to instructional traffic editing and controllable self-driving policy evaluation.
    
\end{enumerate}

\vspace{-0.5em}
\section{Related Work}
\label{sec:related}
\vspace{-0.5em}

\noindent\textbf{Traffic scenario generation} 
traditionally rely on rules defined by human experts \cite{SUMO2018}, \emph{e.g.}, rules that enforce vehicles to stay in lanes, follow the lead vehicles~\cite{trafficmodeler, Real-Time-Traffic, GUNAWAN201267} or change lanes \cite{lane-change}.
This approach is used in most virtual driving datasets \cite{Synscapes, 7780721, Johnson-Roberson:2017aa, Richter2016PlayingFD} and simulators \cite{Dosovitskiy17, SUMO2018, structured_dr}. 
However, traffic scenarios generated in this way often lack realism as they do not necessarily match the distribution of real-world traffic scenarios.
Moreover, creating interesting traffic scenarios in this way requires non-trivial human efforts from experts, making it difficult to scale.
In contrast, \MODELNAME learns the real-world traffic distribution for realistic traffic generation.
Also, \MODELNAME can generate interesting scenarios with language descriptions, largely reducing the requirement of human experts.

Prior work in learning-based traffic generation is more related to our work. SceneGen~\cite{tan2021scenegen} uses autoregressive models to generate traffic scenarios conditioned on ego-vehicle states and maps. TrafficGen~\cite{feng2022trafficgen} applies two separate modules for agent initialization and motion prediction.
BITS~\cite{XuChenEtAl2023} learns to simulate agent motions with a bi-level imitation learning method.
Similar to \MODELNAME, these methods learn to generate realistic traffic scenarios from real-world data.
However, they lack the ability to control traffic generation towards users' preferences. In contrast, \MODELNAME achieves such controllability via natural languages and at the same time can generate highly realistic traffic scenarios.
Moreover, we will show in the experiments that \MODELNAME also outperforms prior work in the setting of unconditional traffic reconstruction, due to our query-based end-to-end architecture.

\noindent\textbf{Text-conditioned generative models} 
have recently shown strong capabilities for controllable content creation for image~\cite{ramesh2021zero}, audio~\cite{huang2023noise2music}, motion~\cite{tevet2022motionclip}, 3D object~\cite{gao2022get3d} and more.
DALL-E~\cite{ramesh2021zero} uses a transformer to model text and image tokens as a single stream of data.
Noise2Music~\cite{huang2023noise2music} uses conditioned diffusion models to generate music clips from text prompts.
MotionCLIP~\cite{tevet2022motionclip} achieves text-to-human-motion generation by aligning the latent space of human motion with pre-trained CLIP~\cite{radford2021learning} embedding.
These methods typically require large-scale pairs of content-text data for training.
Inspired by prior work, \MODELNAME is the first-of-its-kind model for text-conditioned traffic generation. Also, due to the use of LLM and our design of structured representation, \MODELNAME achieves text-conditioned generation \textit{without} any text-traffic paired data.

\noindent\textbf{Large language models} 
have become increasingly popular in natural language processing and related fields due to their ability to generate high-quality text and perform language-related tasks. 
GPT-2~\cite{radford2019language} is a transformer-based language model that is pre-trained on vast amounts of text data. 
GPT-4~\cite{openai2023gpt4} largely improves the instruction following capacity by fine-tuning with human feedback to better align the models with their users~\cite{ouyang2022training}.
In our work, we adapt the GPT-4 model~\cite{openai2023gpt4} with in-context-learning~\cite{min2022rethinking} and chain-of-thought~\cite{wei2022chain} prompting method as our \INTERPRETER.

\vspace{-0.5em}
\section{Preliminaries}
\label{sec:prel}
\vspace{-0.5em}

Let $m$ be a map region, and $\mathbf{s}_t$ be the state of all vehicles in a scene at time $t$.
A traffic scenario $\tau = (m, \mathbf{s}_{1:T})$ is the combination of a map region $m$ and $T$ timesteps of vehicle states $\mathbf{s}_{1:T} = [\mathbf{s}_1, ..., \mathbf{s}_T]$.

\noindent\textbf{Map.}
We represent each map region $m$ by a set of $S$ lane segments denoted by $m=\{v_1, ..., v_S\}$.
Each lane segment includes the start point and endpoint of the lane, the lane type (center lane, edge lane, or boundary lane), and the state of the traffic light control.

\noindent\textbf{Vehicle states.}
The vehicle states $\mathbf{s}_t=\{s_t^1, ..., s_t^N\}$ at time $t$ consist of $N$ vehicle.
For each vehicle, we model the vehicle's position, heading, velocity, and size.
Following prior work~\cite{tan2021scenegen, feng2022trafficgen}, we choose the vehicle at the center of the scenario in the first frame as the ego-vehicle.
It represents the self-driving vehicle in simulation platforms.

\vspace{-0.5em}
\section{\MODELNAME: Language-Conditioned Traffic Generation}
\label{sec:method}
\vspace{-0.5em}

Our goal is to train a language-conditioned model $\tau \sim \text{\MODELNAME}(L, \mathcal{M})$ that produces traffic scenarios from a text description $L$ and a dataset of maps $\mathcal{M}$.
Our model consists of three main components: A language \INTERPRETER (Section~\ref{method:interpreter}) that encodes a text description into a structured representation $z$.
Map \RETRIEVAL $m \sim \text{\RETRIEVAL}(z, \mathcal{M})$ that samples matching map regions $m$ from a dataset of maps $\mathcal{M}$.
A \GENERATOR (Section~\ref{method:generator}) that produces a scenario $\tau \sim \text{\GENERATOR}(z, m)$ from the map $m$ and structured representation $z$.
All components are stochastic, allowing us to sample multiple scenes from a single text description $L$ and map dataset $\mathcal{M}$.
We train the \GENERATOR with a real-world scenario-only driving dataset (Section~\ref{method:train}).
\begin{figure*}[!t]
\centering
% \vspace{-5em}
\includegraphics[width=1.0\linewidth]{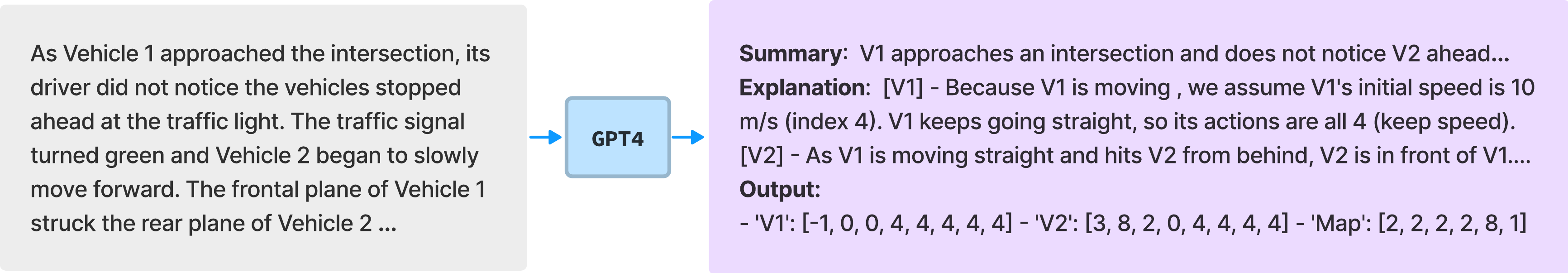}
\vspace{-0.4cm}
\caption{Example \INTERPRETER input and output. We only show partial texts for brevity.}
\vspace{-0.3cm}
\label{fig:gpt}
\end{figure*}

\subsection{\INTERPRETERNORMAL}
\label{method:interpreter}
    The \INTERPRETER takes a text description $L$ as input and produces a structured representation $z = \text{\INTERPRETER}(L)$.
    After defining the representation $z$, we show how to produce it via GPT-4~\cite{openai2023gpt4}.
    
\noindent\textbf{Structured representation} $z = [z^{m}, z^{a}_1, \ldots z^{a}_N]$ contains both map-specific $z^{m}$ and agent-specific components $z^{a}_i$.
For each scenario, we use a $6$-dimensional vector $z^{m}$ describing the local map.
It measures the number of lanes in each direction (north, south, east, west), the distance of the map center to an intersection, and the lane the ego-vehicle finds itself in.
This compact abstract allows a language model to describe the important properties of a $m$ and interact with map dataset $\mathcal{M}$. 
For each agent $i$, $z^a_i$ is an $8$-dimentional integer vector describing the agent relative to the ego vehicle.
It contains an agent's quadrant position index (1-4), distance range (0-20m, 20-40m,...), orientation index (north, south, east, west), speed range (0-2.5m/s, 2.5-5m/s, ...), and action description (turn left, accelerate, ...). 
Please refer to Supp.\ref{supp_interpreter}. for a complete definition of $z$.
Note that the representation $z$ does not have a fixed length, as it depends on the number of agents in a scene.

\noindent\textbf{Language interpretation.}
To obtain the structured representation, we use a large language model (LLM) and formulate the problem into a text-to-text transformation.
Specifically, we ask GPT-4~\cite{openai2023gpt4} to translate the textual description of a traffic scene into a YAML-like description through in-context learning~\cite{min2022rethinking}.
To enhance the quality of the output, we use Chain-of-Thought~\cite{wei2022chain} prompting to let GPT-4 summarize the scenario $q$ in short sentences and plan agent-by-agent how to generate $z$.
See Figure~\ref{fig:gpt} for an example input and output.
Refer to Supp.~\ref{supp_interpreter} for the full prompt and Supp.~\ref{example:input_output} for more complete examples.

\vspace{-0.5em}
\subsection{\RETRIEVALNORMAL}
\vspace{-0.5em}

The \RETRIEVAL module takes a map representation $z^m$ and map dataset $\mathcal{M}$, and samples map regions $m \sim \text{\RETRIEVAL}(z^m, \mathcal{M})$.
Specifically, we preprocess the map dataset $\mathcal{M}$ into potentially overlapping map regions $\{m_1,m_2,...\}$.
We sample map regions, such that their center aligns with the locations of an automated vehicle in an offline driving trace.
This ensures that the map region is both driveable and follows a natural distribution of vehicle locations.
For each map $m_j$, we pre-compute its map representation $\hat z^m_j$.
This is possible, as the map representation is designed to be both easy to produce programmatically and by a language model.
Given $z^m$, the \RETRIEVAL ranks each map region $m_j$ based its feature distance ${\left\lVert z^m - \hat z_j^m \right\rVert}$.
Finally, \RETRIEVAL randomly samples $m$ from the top-$K$ closest map regions.

\vspace{-0.5em}
\subsection{\GENERATORNORMAL}
\vspace{-0.5em}
\label{method:generator}

% \begin{figure*}
% \centering
% % \vspace{-6em}
% \includegraphics[width=1.0\linewidth]{figures/project_figures/training_pipeline.pdf}
% % \vspace{-5em}
% \caption{Model overview of our proposed scenario generation model. \STnote{TODO: update it}}
% \label{fig:method}
% \end{figure*}

\begin{figure*}
\centering
% \vspace{-4em}
\includegraphics[width=1.0\linewidth]{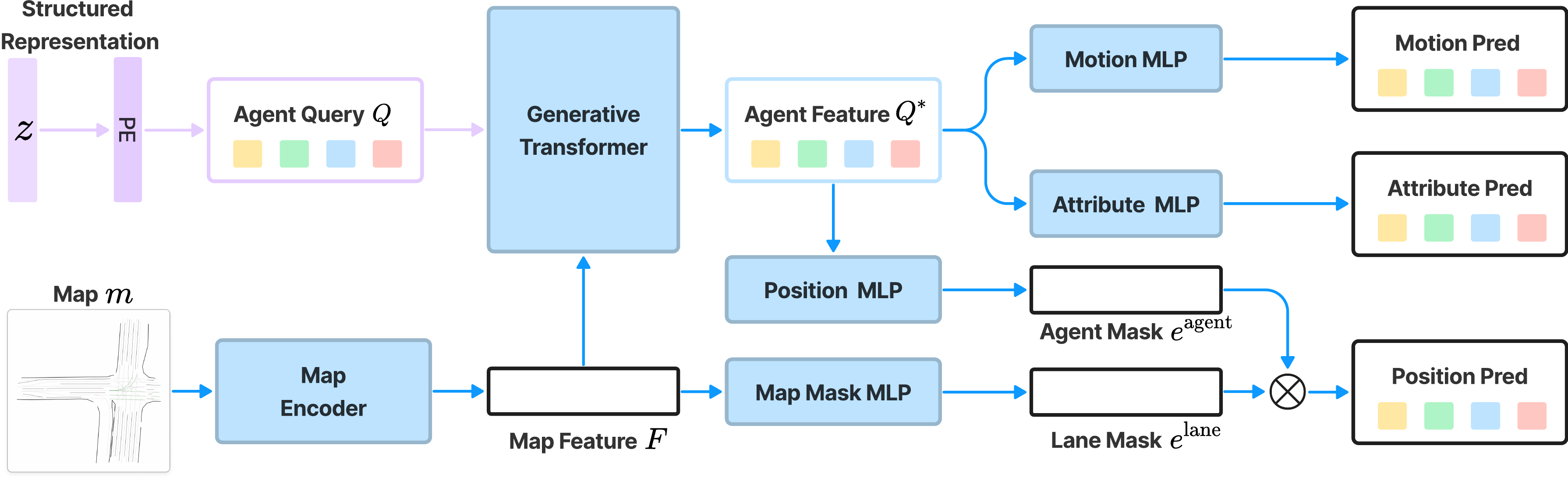}
% \includegraphics[width=1.0\linewidth]{figures/method_figures/generator_vertical.pdf}
% \vspace{-4em}
\caption{Architecture of our \GENERATOR model.}
\vspace{-1.5em}
\label{fig:generator}
\end{figure*}

Given a structured representation $z$ and map $m$, the \GENERATOR produces a traffic scenario $\tau = \text{\GENERATOR}(z, m)$.
We design \GENERATOR as a query-based transformer model to efficiently capture the interactions between different agents and between agents and the map. 
It places all the agents in a single forward pass and supports end-to-end training.
The \GENERATOR has four modules (Figure~\ref{fig:generator}): 1) a map encoder that extracts per-lane map features $F$; 2) an agent query generator that converts structured representation $z^a_i$ to agent query $q_i$; 3) a generative transformer that models agent-agent and agent-map interactions; 4) a scene decoder to output the scenario $\tau$.

\noindent\textbf{Map encoder} processes a map region $m=\{v_1, \ldots, v_S\}$ with $S$ lane segments $v_i$ into a map feature $F = \{f_1,\ldots, f_S\}$, and meanwhile fuse information across different lanes.
Because $S$ could be very large, we use \textit{multi-context gating} (MCG) blocks~\cite{varadarajan2021multipath} for efficient information fusion.
MCG blocks approximate a transformer's cross-attention, but only attend a single global context vector in each layer.
Specifically, a MCG block takes a set of features $v_{1:S}$ as input, computes a context vector $c$, then combines features and context in the output $v^\prime_{1:S}$.
Formally, each block is implemented via
\begin{equation*}
\label{eq:map_encoder}
  v^\prime_i = \text{MLP}(v_i) \odot \text{MLP}(c) \quad \text{where} \quad c = \text{MaxPool}(v_{1:S})
\end{equation*}
where $\odot$ is the element-wise product.
The encoder combines 5 MCG blocks with 2-layer MLPs.

\noindent\textbf{Agent query generator} transforms the structured representation $z^a_i$ of each agent $i$ into an agent query $q_i \in \mathbb{R}^{d}$. 
We implement this module as an MLP of positional embeddings of the structured representation
    $q_i = \text{MLP}(\text{PE}(z^a_i)) + \text{MLP}(x_i).$
We use a sinusoidal position encoding $\text{PE}(\cdot)$.
We also add a learnable query vector $x_i$ as inputs, as inspired by the object query in DETR~\cite{detr}. 

\noindent\textbf{Generative transformer.} To model agent-agent and agent-map interactions, we use $F = \{f_1, \ldots, f_S\}$ and $Q = \{q_1, \ldots, q_N\}$ as inputs and pass them through multiple transformer layers.
Each layer follows
    $Q^\prime = \texttt{MHCA}(\texttt{MHSA}(Q), F)$,
where \texttt{MHCA}, \texttt{MHSA} denote that multi-head cross-attention and multi-head self-attention respectively~\cite{NIPS2017_3f5ee243}. 
The output of $Q'$ in each layer is used as the query for the next layer to cross-attend to $F$. 
The outputs of the last-layer $Q^*$ are the agent features.

\noindent\textbf{Scene decoder}.
For each agent feature $q^*_i$, the scene decoder produces the agents position, attributes, and motion using an MLP. 
To decode the position, we draw inspiration from MaskFormer~\cite{cheng2021maskformer}, placing each actor on a lane segment in the map. 
This allows us to explicitly model the positional relationship between each actor and the road map.
Specifically, we employ an MLP to turn $q^*_i$ into an actor mask embedding $e_i^{\text{agent}} \in \mathbb{R}^{d}$.
Likewise, we transform each lane feature $f_j$ into a per-lane map mask embedding $e_j^{\text{lane}} \in \mathbb{R}^{d}$. 
The position prediction $\hat{p}_i \in \mathbb{R}^S$ for the $i$-th agent is then ${\hat{p}_i = \text{softmax}(e_i^{\text{agent}} \times [e^{\text{lane}}_1, \ldots, e^{\text{lane}}_S]^T)}$,

For each agent query, we predict its \textbf{attributes}, namely heading, velocity, size, and position shift from the lane segment center, following Feng et al.~\cite{feng2022trafficgen}.
The attribute distribution of a potential agent is modeled with a Gaussian mixture model (GMM).
The parameters of a $K$-way GMM for each attribute of agent $i$ are predicted as
$[\mu_{i}, \Sigma_{i}, \pi_{i}] = \text{MLP}(q^*_i)$,
where $\mu_{i}, \Sigma_{i}$ and $\pi_{i}$ denote the mean, covariance matrix, and the categorical weights of the $K$-way GMM model.

We further predict the future $T-1$ step \textbf{motion} of each agent, by outputting $K'$ potential future trajectories for each agent:
    $\{\text{pos}_{i,k}^{2:T}, \text{prob}_{i,k}\}_{k=1}^{K'} =  \text{MLP}(q^*_i)$,
where $\text{pos}_{i,k}^{2:T}$ represents the $k$-th trajectory states for $T-1$ future steps, and $\text{prob}_{i,k}$ is its probability.
Specifically, for each timestamp $t$, $\text{pos}_{i,k}^t=(x,y,\theta)$ contains the agent's position $(x,y)$ and heading $\theta$ at $t$.

During inference, we sample the most probable values from the predicted position, attribute, and motion distributions of each agent query to generate an output agent status through $T$ time stamps $s^i_{1:T}$.
Compiling the output for all agents, we derive the vehicle statuses $\mathbf{s}_{1:T}$.
In conjunction with $m$, the \GENERATOR outputs the final traffic scenario $\tau = (m, \mathbf{s}_{1:T})$.

\vspace{-0.5em}
\subsection{Training}
\vspace{-0.5em}
\label{method:train}
The \GENERATOR is the only component of \MODELNAME that needs to be trained.
We use real-world self-driving datasets, composed of $D$ traffic scenarios $\{\tau_j\}_{j=1}^D$.
For each traffic scene, we use an \ENCODER to produce the latent representation $z$, then train the \GENERATOR to reconstruct the scenario.

\paragraph{\ENCODERNORMAL.}
The \ENCODER takes a traffic scenario $\tau$ and outputs structured agent representation: $z^a = \text{\ENCODER}(\tau)$.
As mentioned in Section~\ref{method:interpreter},  $z^a$ contains compact abstract vectors of each agent $\{z^a_1,...,z^a_N\}$.
For each agent $i$, the \ENCODER extracts from its position, heading, speed, and trajectory from the ground truth scene measurements $\mathbf{s}_{1:T}^i$ in $\tau$, and converts it into $z^a_i$ following a set of predefined rules.
For example, it obtains the quadrant position index with the signs of $(x,y)$ position.
In this way, we can use \ENCODER to automatically convert any scenario $\tau$ to latent codes $z$.
This allows us to obtain a paired dataset $(m, \mathbf{s}_{1:N}, z^{a}_{1:N})$ from scenario-only driving dataset.

\paragraph{Training objective.}
For each data sample $(m, \mathbf{s}_{1:N}, z^{a}_{1:N})$, we generate a prediction $p = \text{\GENERATOR} (z, m)$.
The objective is to reconstruct the real scenario $\tau$. 
We compute the loss as:
\begin{equation}
\label{equ:full_loss}
\mathcal{L}(p, \tau) = \mathcal{L}{_\text{position}}(p, \tau) + \mathcal{L}_{\text{attr}}(p, \tau) + \mathcal{L}_{\text{motion}}(p, \tau),
\end{equation}
where $\mathcal{L}_{\text{position}}, \mathcal{L}_{\text{attr}}, 
\mathcal{L}_{\text{motion}}$ are losses for each of the different predictions.
We pair each agent in $p$ with a ground-truth agent in $\tau$ based on the sequential ordering of the structured agent representation $z^a$.
We then calculate loss values for each component.
For $\mathcal{L}_{\text{position}}$, we use cross-entropy loss between the categorical output $\hat{p}$ and the ground-truth lane segment id.
For $\mathcal{L}_{\text{attr}}$, we use a negative log-likelihood loss, computed using the predicted GMM on the ground-truth attribute values.
For $\mathcal{L}_{\text{motion}}$, we use MSE loss for the predicted trajectory closest to the ground-truth trajectory.
The training objective is the expected loss $\mathcal{L}$ over the dataset.
We refer readers to Supp.~\ref{supp_generator} for more detailed formulations of the loss functions.

\section{Experiments}
\label{sec:exp}
\vspace{-0.1cm}
\noindent\textbf{Datasets.} We use the large-scale real-world Waymo Open Dataset~\cite{waymo_open_dataset}, partitioning it into 68k traffic scenarios for training and 2.5k for testing.
For each scene, we limit the maximum number of lanes to $S=384$, and set the maximum number of vehicles to $N=32$.
We simulate $T=50$ timesteps at 10 fps, making each $\tau$ represent a 5-second traffic scenario.
We collect all the map segments in the Waymo Open Dataset training split for our map dataset $\mathcal{M}$.

\noindent\textbf{Implementation.}
We query GPT-4~\cite{openai2023gpt4} (with a temperature of $0.2$) through the OpenAI API for \INTERPRETER. 
For \GENERATOR, we set the latent dimension $d=256$. 
We use a 5-layer MCG block for the map encoder.
For the generative transformer, we use a 2-layer transformer with 4 heads.
We use a dropout layer after each transformer layer with a dropout rate of $0.1$.
For each attribute prediction network, we use a 2-layer MLP with a latent dimension of $512$.
For attribute GMMs, we use $K=5$ components. For motion prediction, we use $K'=12$ prediction modes.
We train \GENERATOR with AdamW~\cite{Loshchilov2017DecoupledWD} for 100 epochs, with a learning rate of 3e-4 and batch size of $64$.

\vspace{-0.3em}
\subsection{Scene Reconstruction Evaluation}
\vspace{-0.2em}
We evaluate the quality of \MODELNAME's generated scenarios by comparing them to real scenarios from the driving dataset.
For each scenario sample $(\tau, z, m)$ in the test dataset, we generate a scenario with $\hat{\tau} = \text{\GENERATOR} (z,m)$ and then compute different metrics with $\tau$ and $\hat{\tau}$.

\noindent\textbf{Metrics}.
To measure the realism of scene initialization, we follow~\cite{tan2021scenegen, feng2022trafficgen} and compute the maximum mean discrepancy (\textbf{MMD}~\cite{gretton2012kernel}) score for actors' positions, headings, speed and sizes.
Specifically, MMD measures the distance between two distributions $q$ and $p$.
For each pair of real and generated data $(\tau, \hat{\tau})$, we compute the distribution difference between them per attribute.
To measure the realism of generated motion behavior, we employ the standard mean average distance error (\textbf{mADE}) and mean final distance error (\textbf{mFDE}). 
For each pair of real and generated scenarios $(\tau, \hat{\tau})$, we first use the Hungarian algorithm to compute a matching based on agents' initial locations with their ground-truth location. 
We then transform the trajectory for each agent based on its initial position and heading to the origin of its coordinate frame, to obtain its relative trajectory. 
Finally, we compute mADE and mFDE using these relative trajectories. 
We also compute the scenario collision rate (\textbf{SCR}), which is the average proportion of vehicles involved in collisions per scene.

\noindent\textbf{Baselines}. 
We compare against a state-of-the-art traffic generation method, \textbf{TrafficGen}~\cite{feng2022trafficgen}.
As TrafficGen only takes a map $m$ as input to produce a scenario $\tau$, we train a version of \MODELNAME that also only uses $m$ as input for a fair comparison, referred to as \MODELNAME (w/o $z$).
We also compare against \textbf{MotionCLIP}~\cite{tevet2022motionclip}, which takes both a map $m$ and text $L$ as input to generate a scenario $\tau$.
Please refer to Supp.~\ref{supp_exp} for the implementation details of each baseline. 

\noindent\textbf{Results}.
\begin{table*}
\centering
\begin{tabular}{lccccccc}
\toprule
\multirow{2}{*}{Method} & \multicolumn{4}{c}{Initialization} & \multicolumn{3}{c}{Motion} \\
& Pos & Heading & Speed & Size & mADE &  mFDE & SCR  \\\cmidrule{1-8}
TrafficGen~\cite{feng2022trafficgen} & 0.2002 & 0.1524 & 0.2379 & \textbf{0.0951} & 10.448 & 20.646 & \textbf{5.690} \\
MotionCLIP~\cite{tevet2022motionclip} & 0.1236 & 0.1446 & 0.1958 & 0.1234 & 6.683 & 13.421 & 8.842   \\
\MODELNAME (w/o $z$) & 0.1319 & 0.1418 & 0.1948 & 0.1092 & 6.315 & 12.260 & 8.383 \\
\midrule
\MODELNAME & \textbf{0.0616} & \textbf{0.1154} & \textbf{0.0719} & 0.1203 & \textbf{1.329} & \textbf{2.838} & 6.700 \\
\midrule
\end{tabular}
\vspace{-0.2cm}
\caption{Traffic scenario generation realism evaluation (lower the better).
}
\label{tab:fiedelity_main}
\vspace{-0.6cm}
\end{table*}

The results in Table~\ref{tab:fiedelity_main} indicate the superior performance of \MODELNAME. 
In terms of scene initialization, \MODELNAME (w/o $z$) outperforms TrafficGen in terms of MMD values for the Position, Heading, and Speed attributes. 
Importantly, when conditioned on the language input $L$, \MODELNAME significantly improves its prediction of Position, Heading, and Speed attributes, significantly outperforming both TrafficGen and MotionCLIP on MMD ($> 2\times$).
\MODELNAME also achieves 7-8x smaller mADE and mFDE than baselines when comparing generated motions. 
The unconditional version of \MODELNAME, without $z$, also outpaces TrafficGen in most metrics, demonstrating the effectiveness of \GENERATOR's query-based, end-to-end transformer design. 
We note that \MODELNAME (w/o) $z$ has an on-par Size-MMD score with TrafficGen, which is lower than \MODELNAME. We conjecture that this is because our model learns spurious correlations of size and other conditions in $z$ in the real data.

\vspace{-0.5em}
\subsection{Language-conditioned Simulation Evaluation}
\vspace{-0.5em}
\label{sec:exp_human}
\MODELNAME aims to generate a scenario $\tau$ that accurately represents the traffic description from the input text $L$. 
Since no existing real-world text-scenario datasets are available, we carry out our experiment using text $L$ from a text-only traffic scenario dataset. 
To evaluate the degree of alignment between each scenario and the input text, we conduct a human study. 
We visualize the output scenario $\tau$ generated by \MODELNAME or the baselines, and ask humans to assess how well it matches the input text.

\noindent\textbf{Datasets.}
We use a challenging real-world dataset, the \textbf{Crash Report} dataset~\cite{national2016crash}, provided by the NHTSA. 
Each entry in this dataset comprises a comprehensive text description of a crash scenario, including the vehicle's condition, driver status, road condition, vehicle motion, interactions, and more.
Given the complexity and intricate nature of the traffic scenarios and their text descriptions, this dataset presents a significant challenge (see Figure~\ref{fig:gpt} for an example).
We selected 38 cases from this dataset for the purposes of our study.
For a more controllable evaluation, we also use an \textbf{Attribute Description} dataset.
This dataset comprises text descriptions that highlight various attributes of a traffic scenario. 
These include aspects like sparsity ("the scenario is dense"), position ("there are vehicles on the left"), speed ("most cars are driving fast"), and the ego vehicle's motion ("the ego vehicle turns left"). 
We create more complex descriptions by combining 2, 3, and 4 attributes. 
This dataset includes 40 such cases.
Refer to Supp.~\ref{supp_exp} for more details about these datasets.

\noindent\textbf{Baselines.}
We compare with TrafficGen and MotionCLIP.
For each text input $L$, \MODELNAME outputs a scenario $\tau=(m, \mathbf{s}_{1:T})$.
To ensure fairness, we feed data $m$ to both TrafficGen and MotionCLIP to generate scenarios on the same map. 
As TrafficGen does not take language condition as input, we only feed $L$ to MotionCLIP.
In addition, TrafficGen can't automatically decide the number of agents, therefore it uses the same number of agents as our output $\tau$.

\noindent\textbf{Human study protocol}. For each dataset, we conduct a human A/B test. 
We present the evaluators with a text input, along with a pair of scenarios generated by two different methods using the same text input, displayed in a random order. 
The evaluators are then asked to decide which scenario they think better matches the text input. 
Additionally, evaluators are requested to assign a score between 1 and 5 to each generated scenario, indicating its alignment with the text description; a higher score indicates a better match. 
A total of 12 evaluators participated in this study, collectively contributing 1872 scores for each model.

\noindent\textbf{Quantitative Results}.
\begin{table*}[t]
\centering
\begin{tabular}{lccccccc}
\toprule
\multirow{2}{*}{Method} & \multicolumn{2}{c}{Crash Report} & \multicolumn{2}{c}{Attribute Description} \\
& Ours Prefered (\%) & Score (1-5) & Ours Prefered (\%) & Score (1-5) \\\cmidrule{1-5}
TrafficGen~\cite{feng2022trafficgen} & 92.35 & 1.58 & 90.48 & 2.43 \\
MotionCLIP~\cite{tevet2022motionclip} & 95.29 & 1.65 & 95.60 & 2.10 \\
\midrule
\MODELNAME & - & \textbf{3.86} & - & \textbf{4.29} \\
\midrule
\end{tabular}
\vspace{-0.2cm}
\caption{Human study results on the language-conditioned simulation.}
\vspace{-0.7em}
\label{tab:human_study}
\end{table*}

We show the results in Table~\ref{tab:human_study}.
We provide preference score, reflecting the frequency with which \MODELNAME's output is chosen as a better match than each baseline. 
We also provide the average matching score, indicating the extent to which evaluators believe the generated scenario matches the text input.
With \MODELNAME often chosen as the preferred model by human evaluators (\textit{at least} $90\%$ of the time), and consistently achieving higher scores compared to other methods, these results underline its superior performance in terms of text-controllability over previous works. 
The high matching score also signifies \MODELNAME's exceptional ability to generate scenarios that faithfully follow the input text.
We include more analysis of human study result in Supp.~\ref{supp_result_human_study}.

\begin{figure*}
\centering
% \vspace{-4em}
% \includegraphics[width=1.0\linewidth]{figures/result_figures/text_quali.pdf}
\includegraphics[width=0.95\linewidth]{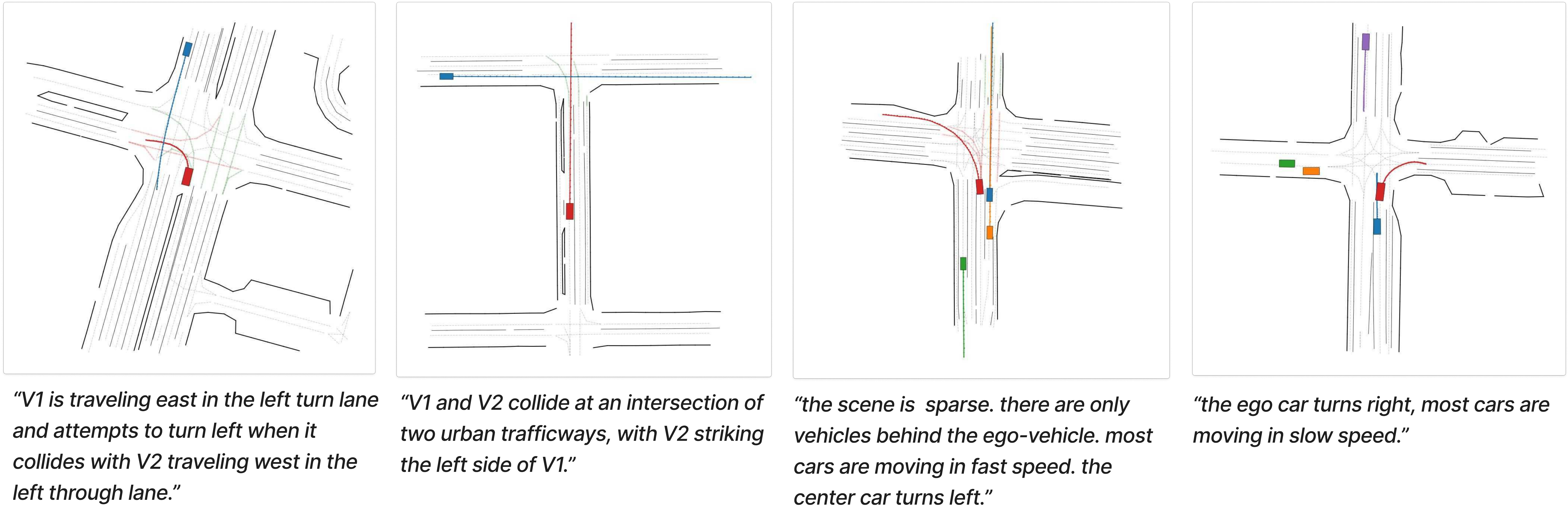}
% \vspace{-4em}
\caption{Qualitative results on text-conditioned generation.}
\vspace{-1.5em}
\label{fig:qualitative}
\end{figure*}

\noindent\textbf{Qualitative Results}.
We show examples of \MODELNAME output given texts from the Crash Report (left two) and Attribute Description (right two) datasets in Figure~\ref{fig:qualitative}.
Each example is a pair of input text and the generated scenario.
Because texts in Crash Report are excessively long, we only show the output summary of our \INTERPRETER for each example (Full texts in Supp.~\ref{supp_exp}).
Please refer to Supp. video for the animated version of the examples here.
We show more examples in Supp.~\ref{supp_result_qualitative}.

\vspace{-0.5em}
\subsection{Application: Instructional Traffic Scenario Editing}
\vspace{-0.5em}
\label{sec:exp_edit}
\begin{figure*}
\centering
% \vspace{-4em}
\includegraphics[width=1.0\linewidth]{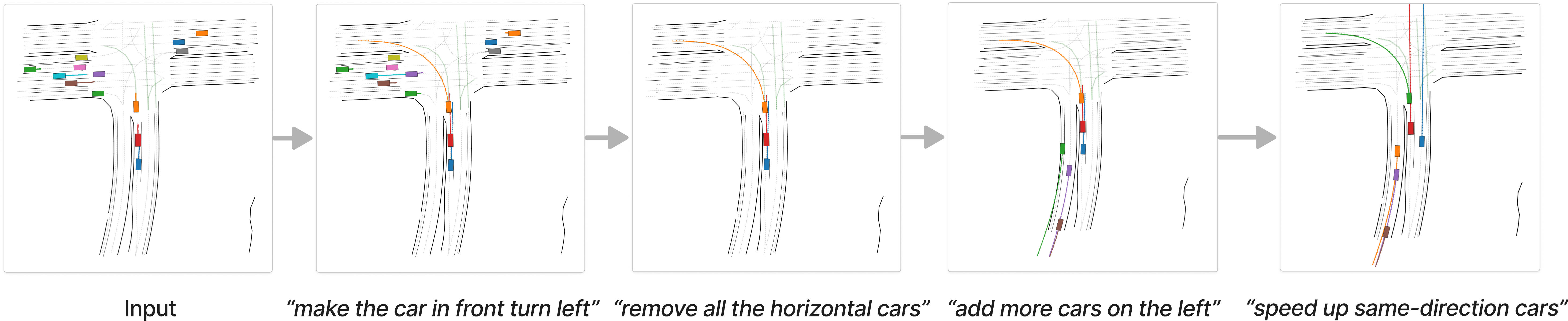}

% \vspace{-4em}
\caption{Instructional editing on a real-world scenario. Refer to Supp.A for full prompts.}
% \vspace{-1.5em}
\label{fig:edit}
\end{figure*}
Besides language-conditioned scenario generation, \MODELNAME can also be applied to instructional traffic scenario editing. Given either a real or generated traffic scenario $\tau$, along with an editing instruction text $I$, \MODELNAME can produce an edited scenario $\hat{\tau}$ that follows $I$.
First, we acquire the structured representation of the scenario using $z = \text{\ENCODER}(\tau)$. Next, we compose a unique prompt that instructs \INTERPRETER to alter $z$ in accordance with $I$, resulting in $\hat{z} = \text{\INTERPRETER}(z, I)$. Finally, we generate the edited scenario $\hat{\tau} = \text{\GENERATOR}(\hat{z}, m)$, where $m$ is the same map used in the input.

We show an example of consecutive instructional editing of a real-world scenario in Figure~\ref{fig:edit}. 
We can see that \MODELNAME supports high-level editing instructions (vehicle removal, addition and action change). It produces realistic output following the instruction.
This experiment highlights \MODELNAME's potential for efficient instruction-based traffic scenario editing.
As another application of \MODELNAME, we also show how \MODELNAME can be utilized to generate interesting scenarios for controllable self-driving policy evaluation. Please refer to Supp.~\ref{supp_result_policy} for this application.

\vspace{-0.5em}
\subsection{Ablation study}
\begin{table*}
% \vspace{-0.5em}
\centering
\begin{subtable}[t]{0.45\linewidth}
\resizebox{\textwidth}{!}{
\centering
\begin{tabular}{lcccccc}\toprule
Method & Pos & Heading & Speed & Size \\\cmidrule{1-5}
w/o Quad. & 0.092 & 0.122 & 0.076 & 0.124 \\
w/o Dist. & 0.071 & 0.124 & 0.073 & 0.121 \\
w/o Ori. & 0.067 & 0.132 & 0.082 & 0.122 \\
\midrule
\MODELNAME & \textbf{0.062} & \textbf{0.115} & \textbf{0.072} & \textbf{0.120} \\
\midrule
\end{tabular}}
\vspace{-0.2cm}
\caption{Ablation study for scene initialization.}
\label{tab:table1}
\end{subtable}%
\vspace{0.1cm}
\begin{subtable}[t]{0.5\linewidth}
\resizebox{\textwidth}{!}{
\centering
\begin{tabular}{lcccccc}\toprule
Method & mADE & mFDE & SCR \\\cmidrule{1-4}
w/o Speed & 2.611 & 5.188 & 7.150 \\
w/o Action & 2.188 & 5.099 & 7.416 \\
\MODELNAME init. + \cite{feng2022trafficgen} motion & 2.467 & 5.682 & \textbf{5.210} \\\midrule
\MODELNAME & \textbf{1.329} & \textbf{2.838} & 6.700 \\
\midrule
\end{tabular}}
\vspace{-0.2cm}
\caption{Ablation study for motion behavior generation.}
\end{subtable}
\vspace{-0.2cm}
\caption{Scene reconstruction ablation study on the Waymo Open Dataset.}
\label{tab:abl}
\vspace{-0.4cm}
\end{table*}

\vspace{-0.5em}

\noindent\textbf{Scene initialization.} Table~\ref{tab:abl} summarizes the results, where the last row corresponds to our full method. 
To validate the performance of \MODELNAME for scene initialization, we mask out the quadrant index, distance, and orientation in the structure representation $z$ for each agent, respectively. 
As a result, we observed a significant performance drop, especially in the prediction of Position and Heading attributes, as shown in the left side of Table~\ref{tab:abl}. 
This suggests that including quadrant index, distance, and orientation in our structured representation is effective.

\noindent\textbf{Motion behavior generation.} We summarized the results in Table~\ref{tab:abl} (right). By masking out the speed range and action description in the structured representation for each agent, we observed a significant performance drop in the metrics for motion behavior. 
Moreover, if we initialize the scene with \MODELNAME while generating agents' motion behavior using TrafficGen's~\cite{feng2022trafficgen}, we also observed significantly worse performance than using \MODELNAME to generate the traffic scenario in one shot. The results suggest that the end-to-end design of scene initialization and motion behavior generation by our \MODELNAME can lead to better performance.
We show more ablation study results in Supp.~\ref{supp_result_ablation}.

\section{Conclusion}
\label{sec:conclusion}
\vspace{-0.5em}
In this work, we present \MODELNAME, a first-of-its-kind method for language-conditioned traffic scene generation. 
By harnessing the expressive power of natural language, \MODELNAME can generate realistic and interesting traffic scenarios.
The realism of our generated traffic scenes notably exceeds previous state-of-the-art methods.
We further show that \MODELNAME can be applied to applications such as instructional traffic scenario editing and controllable driving policy evaluation.

\noindent\textbf{Limitations.} The primary constraint of \MODELNAME lies in the \INTERPRETER module's inability to output perfect agent placements and trajectories,
as it lacks direct access to detailed lane information from the map.
Our future work aims to overcome these issues by equipping the \INTERPRETER with map and math APIs, enabling it to fetch precise map data and output more comprehensive traffic scenarios.

\noindent\textbf{Acknowledgement.} We thank Yuxiao Chen, Yulong Cao, and Danfei Xu for their insightful discussions. This material is supported by the National Science Foundation under Grant No. IIS-1845485.
\bibliography{egbib}

\newpage
\appendix

% ---- pseudo code coloring ----
\definecolor{codegreen}{rgb}{0,0.6,0}
\definecolor{codegray}{rgb}{0.5,0.5,0.5}
\definecolor{codepink}{RGB}{252, 142, 172}
\definecolor{codepurple}{rgb}{0.58,0,0.82}
% \definecolor{backcolour}{rgb}{0.95,0.95,0.92}
\definecolor{backcolour}{RGB}{245,245,245}
\lstdefinestyle{mystyle}{
    backgroundcolor=\color{backcolour},   
    commentstyle=\color{magenta},
    keywordstyle=\color{blue},
    numberstyle=\tiny\color{codegray},
    stringstyle=\color{codepurple},
    basicstyle=\fontfamily{\ttdefault}\footnotesize,
    breakatwhitespace=false,         
    breaklines=true,                 
    % captionpos=b,                    
    keepspaces=true,    
    frame=single,
    % numbers=left,                    
    numbersep=5pt,                  
    showspaces=false,                
    showstringspaces=false,
    showtabs=false,                  
    tabsize=2,
    classoffset=1, % starting new class
    % otherkeywords={range},
    keywordstyle=\color{violet},
    classoffset=0,
}
\lstset{style=mystyle}

% Define custom prompt style
\lstdefinestyle{prompt}{
    language=sh,
    basicstyle=\footnotesize\ttfamily\color{white},
    keywordstyle=\color{white},
    emphstyle=\color{black},
    commentstyle=\color{green},
    showstringspaces=false,
    numbers=left,
    numberstyle=\footnotesize\color{white},
    frame=single,
    framesep=5pt,
    rulecolor=\color{black},
    xleftmargin=15pt,
    framexleftmargin=10pt,
    backgroundcolor=\color{black},
    moredelim=[il][\textcolor{black}]{$ $},
    moredelim=[is][\textcolor{black}]{\%\%}{\%\%},
}

\renewcommand\thesection{\Alph{section}} 
\setcounter{table}{0}
\renewcommand{\thetable}{A\arabic{table}}
\setcounter{figure}{0}
\renewcommand{\thefigure}{A\arabic{figure}}

\renewcommand{\baselinestretch}{0.9}

\renewcommand\lstlistingname{Prompt}

\section*{Appendix}
In the appendix, we provide implementation and experiment details of our method as well as additional results.
In Section~\ref{supp_interpreter} and Section~\ref{supp_generator}, we show details of \INTERPRETER and \GENERATOR respectively.
In Section~\ref{supp_exp} we present implementation details of our experiments.
Finally, in Section~\ref{supp_result}, we show more results on applications ablation study, as well as additional qualitative results.

\section{\INTERPRETERNORMAL}
\label{supp_interpreter}

\subsection{Structured representation details}

The map specific $z^m$ is a 6-dim integer vector. 
Its first four dimensions denote the number of lanes in each direction (set as north for the ego vehicle). 
The fifth dimension represents the discretized distance in 5-meter intervals from the map center to the nearest intersection (0-5, 5-10...). 
The sixth dimension indicates the ego vehicle's lane id, starting from 1 for the rightmost lane.

For agent $i$, the agent-specific $z_i^a$ is an 8-dim integer vector describing this agent relative to the ego vehicle.
The first dimension denotes the quadrant index (1-4), where quadrant 1 represents the front-right of the ego vehicle. 
The second dimension is the discretized distance to the ego vehicle with a 20m interval, and the third denotes orientation (north, south, east, west). 
The fourth dimension indicates discretized speed, set in 2.5m/s intervals. 
The last four dimensions describe actions over the next four seconds (one per second) chosen from a discretized set of seven possible actions: lane changes (left/right), turns (left/right), moving forward, accelerating, decelerating, and stopping.

\subsection{Generation prompts}

The scenario generation prompt used for \INTERPRETER consists of several sections:

\begin{enumerate}
    \item \textbf{Task description}: simple description of task of scenario generation and output formats.
    \item \textbf{Chain-of-thought prompting~\cite{wei2022chain}}:  For example, "summarize the scenario in short sentences", "explain for each group of vehicles why they are put into the scenario".
    \item \textbf{Description of structured representation}: detailed description for each dimension of the structured representation. We separately inform the model Map and Actor formats.
    \item \textbf{Guidelines}: several generation instructions. For example, "Focus on realistic action generation of the motion to reconstruct the query scenario".
    \item \textbf{Few-shot examples}: A few input-output examples. We provide a Crash Report example.
\end{enumerate}

We show the full prompt below:

\renewcommand\lstlistingname{Prompt}
\begin{lstlisting}[breaklines=true,caption={Full prompt for \INTERPRETER scenario generation.}]
You are a very faithful format converter that translate natrual language traffic scenario descriptions to a fix-form format to appropriately describe the scenario with motion action. You also need to output an appropriate map description that is able to support this scenario. Your ultimate goal is to generate realistic traffic scenarios that faithfully represents natural language descriptions normal scenes that follows the traffic rule.

Answer with a list of vectors describing the attributes of each of the vehicles in the scenario.

Desired format:
Summary: summarize the scenario in short sentences, including the number of vehicles. Also explain the underlying map description.
Explaination: explain for each group of vehicles why they are put into the scenario and how they fullfill the requirement in the description.
Actor Vector: A list of vectors describing the attributes of each of the vehicles in the scenario, only output the values without any text:
- 'V1': [,,,,,,,]
- 'V2': [,,,,,,,]
- 'V3': [,,,,,,,]
Map Vector: A vector describing the map attributes, only output the values without any text:
- 'Map': [,,,,,]

Meaning of the Actor vector attribute:
- dim 0: 'pos': [-1,3] - whether the vehicle is in the four quadrant of ego vechile in the order of [0 - 'front left', 1 - 'back left', 2- 'back right', 3 - 'front right']. -1 if the vehicle is the ego vehicle.
- dim 1: 'distance': [0,3] - the distance range index of the vehicle towards the ego vehicle; range is from 0 to 72 meters with 20 meters interval. 0 if the vehicle is the ego vehicle. For example, if distance value is 15 meters, then the distance range index is 0.
- dim 2: 'direction': [0,3] - the direction of the vehicle relative to the ego vehicle, in the order of [0- 'parallel_same', 1-'parallel_opposite', 2-'perpendicular_up', 3-'perpendicular_down']. 0 if the vehicle is the ego vehicle.
- dim 3: 'speed': [0,20] - the speed range index of the vehicle; range is from 0 to 20 m/s with 2.5 m/s interval. For example, 20m/s is in range 8, therefore the speed value is 8.
- dim 4-7: 'action': [0,7] - 4-dim, generate actions into the future 4 second with each two actions have a time interval of 1s (4 actions in total), the action ids are [0 - 'stop', 1 - 'turn left', 2 - 'left lane change', 3- 'decelerate', 4- 'keep_speed', 5-'accelerate',  6-'right lane change', 7-'turn right'].

Meaning of the Map attributes:
- dim 0-1: 'parallel_lane_cnt': 2-dim. The first dim is the number of parallel same-direction lanes of the ego lane, and the second dim is the number of parallel opposite-direction lanes of the ego lane.
- dim 2-3: 'perpendicular_lane_cnt': 2-dim. The first dim is the number of perpendicular upstream-direction lanes, and the second dim is the number of perpendicular downstream-direction lanes.
- dim 4: 'dist_to_intersection': 1-dim. the distance range index of the ego vehicle to the intersection center in the x direction, range is from 0 to 72 meters with 5 meters interval. -1 if there is no intersection in the scenario.
- dim 5: 'lane id': 1-dim. the lane id of the ego vehicle, counting from the rightmost lane of the same-direction lanes, starting from 1. For example, if the ego vehicle is in the rightmost lane, then the lane id is 1; if the ego vehicle is in the leftmost lane, then the lane id is the number of the same-direction lanes.

Transform the query sentence to the Actor Vector strictly following the rules below:
- Focus on realistic action generation of the motion to reconstruct the query scenario.
- Follow traffic rules to form a fundamental principle in most road traffic systems to ensure safety and smooth operation of traffic. You should incorporate this rule into the behavior of our virtual agents (vehicles).
- Traffic rule: in an intersection, when the vehicles on one side of the intersection are crossing, the vehicles on the other side of the intersection should be waiting. For example, if V1 is crossing the intersection and V2 is on the perpendicular lane, then V2 should be waiting.
- For speed and distance, convert the unit to m/s and meter, and then find the interval index in the given range.
- Make sure the position and direction of the generated vehicles are correct.
- Describe the initialization status of the scenario.
- During generation, the number of the vehicles is within the range of [1, 32].
- Always generate the ego vehicle first (V1).
- Always assume the ego car is in the center of the scene and is driving in the positive x direction.
- In the input descriptions, regard V1, Vehicle 1 or Unit #1 as the ego vehicle. All the other vehicles are the surrounding vehicles. For example, for "Vehicle 1 was traveling southbound", the ego car is Vehicle 1.
- If the vehicle is stopping, its speed should be 0m/s (index 0). Also, if the first action is 'stop', then the speed should be 0m/s (index 0).
- Focus on the interactions between the vehicles in the scenario.
- Regard the last time stamp as the time stamp of 5 second into the future.

Generate the Map Vector following the rules below:
- If there is vehicle turning left or right, there must be an intersection ahead.
- Should at least have one lane with the same-direction as the ego lane; i.e., the first dim of Map should be at least 1. For example, if this is a one way two lane road, then the first dim of Map should be 2.
- Regard the lane at the center of the scene as the ego lane.
- Consider the ego car's direction as the positive x direction. For example, for "V1 was traveling northbound in lane five of a five lane controlled access roadway", there should be 5 lanes in the same direction as the ego lane.
- The generated map should strictly follow the map descriptions in the query text. For example, for "Vehicle 1 was traveling southbound", the ego car should be in the southbound lane.
- If there is an intersection, there should be at least one lane in either the upstream or downstream direction. 
- If there is no intersection, the distance to the intersection should be -1.
- There should be vehicle driving vertical to the ego vehicle in the scene only when there is an intersection in the scene. For example, when the road is just two-way, there should not be any vehicle driving vertical to the ego vehicle.
- If no intersection is mentioned, generate intersection scenario randomly with real-world statistics.


Query: The crash occurred during daylight hours on a dry, bituminous, two-lane roadway under clear skies.  There was one northbound travel lane and one southbound travel lane with speed limit of 40 km/h (25 mph).   The northbound lane had a -3.6 percent grade and the southbound lane had a +3.6 percent grade.  Both travel lanes were divided by a double yellow line. A 2016 Mazda CX-3 (V1) was in a parking lot attempting to execute a left turn to travel south.  A 2011 Dodge Charger (V2/police car) was traveling north responding to an emergency call with lights sirens activated. V1 was in a parking lot (facing west) and attempted to enter the roadway intending to turn left.   As V1 entered the roadway it was impacted on the left side by the front of V2 (Event 1).  V1 then rotated counterclockwise and traveled off the west road edge and impacted an embankment with its front left bumper (Event 2).  After initial impact V2 continued on in a northern direction and traveling to final rest approximately 40 meters north of impact area facing north in the middle of the roadway.  V1 and V2 were towed from the scene due to damage.

Summary: V1 attempts to turn left from a parking lot onto a two-lane roadway and is struck by V2, a police car traveling north with lights and sirens activated. There are 2 vehicles in this scenario. This happens on a parking lot to a two-lane two-way road with intersection.
Explanation:
- V1 (ego vehicle) is attempting to turn left from a parking lot onto the roadway. We cannot find V1's speed in the query. Because V1 tries to turn left, its initial speed should be set low. We set V1's speed as 5 m/s, which has the index of 2. V1 turns left, so its actions are all 1 (turn left).
- V2 is a police car traveling north with lights and sirens activated. As V1 is turning left, 5 seconds before the crash, V1 is facing west and V2 is coming from northbound, crossing the path of V1. In the coordinates of V1 (which is facing west initially), V2 comes from the front and is on the left side. Hence, V2's position is "front left" (3). As V1 is facing west and V2 facing north, V2 is moving in the perpendicular down direction with V1. Therefore its direction is 3 (perpendicular_down). We cannot find V2's speed in the query. Because V2 is a police car responding to an emergency call, we assume V2's init speed is 10 m/s (index 4). Given this speed, V2's distance to V1 is 10m/s * 5s = 50m (index 10). V2 keeps going straight, so its actions are all 4 (keep speed).
- Map: V1 tries to turn left from a partking lot onto a two-lane roadway. There are a one-way exit lane from parking lot (one same-direction parallel) and the ego vehicle is in the left turn lane with lane id 1. On the perpendicular side there is a two-lane roadway. V1 is about to turn left, so the distance to the intersection is set to be 10m (index 2).
Actor Vector:
- 'V1': [-1, 0, 0, 2, 1, 1, 1, 1]
- 'V2': [0, 10, 3, 4, 4, 4, 4, 4]
Map Vector:
- 'Map': [1, 0, 1, 1, 2, 1]

Query: INSERT_QUERY_HERE

Output:
\end{lstlisting}

\subsection{Instructional editing prompts}
We also provide \INTERPRETER another prompt for instructional scenario editing. This prompt follow a similar structure to the generation prompt. 
We mainly adopt the task description, guidelines, and examples to scenario editing tasks.
Note that for the instructional editing task, we change the distance interval (second dimension) of agent-specific $z_i^a$ from 20 meters to 5 meters. This is to ensure the unedited agents stay in the same region before and after editing.

We show the full prompt below:

\begin{lstlisting}[breaklines=true,caption={Full prompt for \INTERPRETER instructional scenario editing.}]
You are a traffic scenario editor that edit fix-form traffic scenario descriptions according to the user's natural language instructions.

The user will input a fix-form traffic scenario description as well as the map description. The user also a natural language instruction to modify the scenario. You need to output a fix-form traffic scenario that is modified according to the instruction.

Input format:
- V1: [,,,,,,,]
- V2: [,,,,,,,]
- V3: [,,,,,,,]
- Map: [,,,,,]
Instruction: natural language instruction to modify the scenario.

Output format:
Summary: summarize the scenario in short sentences. summarize the user instruction, and indicate which part of the scenario should be modified.
Explaination: explain step-by-step how each part of the scenario is modified.
Actor Vector: A list of vectors describing the attributes of each of the vehicles. Only the vehicles that are modified should be included in the output.
- V2: [,,,,,,,]

Meaning of the Actor vector attribute:
- dim 0: 'pos': [-1,3] - whether the vehicle is in the four quadrant of ego vechile in the order of [0 - 'front left', 1 - 'back left', 2- 'back right', 3 - 'front right']. -1 if the vehicle is the ego vehicle.
- dim 1: 'distance': [0,14] - the distance range index of the vehicle towards the ego vehicle; range is from 0 to 72 meters with 5 meters interval. 0 if the vehicle is the ego vehicle.
- dim 2: 'direction': [0,3] - the direction of the vehicle relative to the ego vehicle, in the order of [0- 'parallel_same', 1-'parallel_opposite', 2-'perpendicular_up', 3-'perpendicular_down']. 0 if the vehicle is the ego vehicle.
- dim 3: 'speed': [0,8] - the speed range index of the vehicle; range is from 0 to 20 m/s with 2.5 m/s interval. For example, 20m/s is in range 8, therefore the speed value is 8.
- dim 4-7: 'action': [0,7] - 4-dim, generate actions into the future 4 second with each two actions have a time interval of 1s (4 actions in total), the action ids are [0 - 'stop', 1 - 'turn left', 2 - 'left lane change', 3- 'decelerate', 4- 'keep_speed', 5-'accelerate',  6-'right lane change', 7-'turn right'].

Meaning of the Map attributes:
- dim 0-1: 'parallel_lane_cnt': 2-dim. The first dim is the number of parallel same-direction lanes of the ego lane, and the second dim is the number of parallel opposite-direction lanes of the ego lane.
- dim 2-3: 'perpendicular_lane_cnt': 2-dim. The first dim is the number of perpendicular upstream-direction lanes, and the second dim is the number of perpendicular downstream-direction lanes.
- dim 4: 'dist_to_intersection': 1-dim. the distance range index of the ego vehicle to the intersection center in the x direction, range is from 0 to 72 meters with 5 meters interval. -1 if there is no intersection in the scenario.
- dim 5: 'lane id': 1-dim. the lane id of the ego vehicle, counting from the rightmost lane of the same-direction lanes, starting from 1. For example, if the ego vehicle is in the rightmost lane, then the lane id is 1; if the ego vehicle is in the leftmost lane, then the lane id is the number of the same-direction lanes.

Follow the instructions below:
- 'V1' is the ego vehicle, and the other vehicles are the surrounding vehicles.
- The user will input a fix-form traffic scenario description as well as the map description. The user also an natural language instruction to modify the scenario. You need to output a fix-form traffic scenario that is modified according to the instruction.
- First figure out which part of the scenario should be modified according to the instruction. For example, if the instruction is "the vehicle in front of me should turn left", then the vehicle in front of the ego vehicle should be modified.

Input:
Actor vector:
- V1: [-1,  0,  0,  0,  4,  4,  4,  4]
- V2: [ 2,  1,  0,  1,  4,  4,  4,  4]
- V3: [ 3,  3,  0,  1,  4,  4,  4,  0]
- V4: [ 3,  4,  0,  8,  4,  4,  2,  0]
- V5: [ 0,  9,  1,  8, -1,  4,  5, -1]
- V6: [ 3,  5,  0,  0,  0,  0,  0,  0]
- V7: [ 0,  9,  3,  0,  0,  0,  0,  0]
- V8: [ 3, 10,  3,  3,  4,  5,  1,  0]
- V9: [ 0, 10,  3,  0,  0,  0,  0, -1]
- V10: [ 3, 10,  2,  0,  0,  0,  0, -1]
- V11: [ 3, 11,  2,  0,  0,  0,  0,  0]
- V12: [ 3, 11,  2,  0,  0,  7,  0,  0]
- Map: [4, 3, 2, 3, 6, 4]

Instruction: move the vehicle behind the ego vehicle to the opposite lane and move faster.

Output:
Summary: The instruction is to move the vehicle behind the ego vehicle to the opposite lane and accelerate. First find which vehicle is behind the ego vehicle. There are only 1 vechile behind the ego vehicle, that is V2 (with position=2, indicating on the right back side of the ego vehicle). Therefore, the vehicle V2 should be modified.
Explaination: The vehicle V2 is modified to move to the opposite lane and accelerate. The vehicle V2 is in the right back side of the ego vehicle, and the ego vehicle is in the rightmost lane of the same-direction lanes. Therefore, the vehicle V2 should move to the leftmost lane of the opposite-direction lanes. Therefore, V2's direction should be opposite to the ego vehicle, changed to 1 (parallel_opposite). In this lane, V2 should be moved to the left back of the ego car, its position should be changed to 1. V2 should move faster, its speed should be changed to 10 (25 m/s).
Actor vector:
- V2: [ 1,  1,  1, 10,  4,  4,  4,  4]

Instruction: remove all the vehicles on the front of the ego car and moving in the same direction.

Output:
Summary: The instruction is to remove all the vehicles on the front of the ego car and moving in the same direction. First find which vehicles are on the front of the ego vehicle. V3-V12 are all on the front of the ego vehicle. Then, only V3, V4 and V6 has the same direction as the ego vehicle (0). Therefore, V3, V4 and V6 should be removed.
Explaination: V3, V4, V6 are on the front of the ego vehicle and moving in the same direction. V3, V4 and V6 are removed from the scenario.

Actor vector:
- V3: removed.
- V4: removed.
- V6: removed.

Input: INSERT_QUERY_HERE

Output:
\end{lstlisting}

\section{\GENERATORNORMAL}
\label{supp_generator}

\subsection{Training objectives}
In the main paper, we show the full training objective of \GENERATOR as:
\begin{equation}
\label{equ:loss_ours}
\mathcal{L}(p, \tau) = \mathcal{L}{_\text{position}}(p, \tau) + \mathcal{L}_{\text{attr}}(p, \tau) + \mathcal{L}_{\text{motion}}(p, \tau).
\end{equation}
In this section, we provide details of each loss function.
We first pair each agent $\hat{a}_i$ in $p$ with a ground-truth agent $a_i$ in $\tau$ based on the sequential ordering of the structured agent representation $z^a$. Assume there are in total $N$ agents in the scenario.

For $\mathcal{L}_{\text{position}}$, we use cross-entropy loss between the per-lane categorical output $\hat{p}$ and the ground-truth lane segment id $l$. Specifically, we compute it as
% 
% % 
\begin{equation}
    \mathcal{L}{_\text{position}}(p, \tau) = \sum_{i=1}^N -\log \hat{p}_i (l_i),
\end{equation}
where $l_i$ is the index of the lane segment that the $i$-th ground-truth agent $a_i$ is on.

For $\mathcal{L}_{\text{attr}}$, we use a negative log-likelihood loss, computed using the predicted GMM on the ground-truth attribute values.
Recall that for each attribute of agent $i$, we use an MLP to predict the parameters of a GMM model $[\mu_{i}, \Sigma_{i}, \pi_{i}]$. 
Here, we use these parameters to construct a GMM model and compute the likelihood of ground-truth attribute values.
Specifically, we have
\begin{equation}
    \begin{aligned}
    \mathcal{L}{_\text{attr}}(p, \tau) = \sum_{i=1}^N ( &-\log \text{GMM}_{\text{heading,i}} (h_i) - \log \text{GMM}_{\text{vel,i}} (vel_i)
    \\ &-\log \text{GMM}_{\text{size,i}} (bbox_i) -\log \text{GMM}_{\text{pos,i}} (pos_i)),
    \end{aligned}
\end{equation}

where $\text{GMM}_{\text{heading,i}}, \text{GMM}_{\text{vel,i}}, \text{GMM}_{\text{size,i}}, \text{GMM}_{\text{pos,i}}$ represent the likelihood function of the predicted GMM models of agent $i$'s heading, velocity, size and position shift. 
These likelihood values are computed using the predicted GMM parameters.
Meanwhile, $h_i$, $vel_i$, $bbox_i$ and $pos_i$ represent the heading, velocity, size and position shift of the ground-truth agent $a_i$ respectively.

For $\mathcal{L}_{\text{motion}}$, we use MSE loss for the predicted trajectory closest to the ground-truth trajectory following the multi-path motion prediction idea~\cite{varadarajan2021multipath}.
Recall that for each agent $\hat{a}_i$, we predict $K'$ different future trajectories and their probabilities as $\{\text{pos}_{i,k}^{2:T}, \text{prob}_{i,k}\}_{k=1}^{K'} =  \text{MLP}(q^*_i)$. For each timestamp $t$, $\text{pos}_{i,k}^t$ contains the agent's position and heading.
We assume the trajectory of ground-truth agent $a_i$ is $\text{pos}_{i, *}^{2:T}$.
We can compute the index $k^*$ of the closest trajectory from the $K'$ predictions as
$k^* = \argmin_k \sum_{t=2}^T (\text{pos}_{i,k}^t - \text{pos}_{i, *}^t)^2$.
Then, we compute the motion loss for agent $i$ as:
\begin{equation}
    \mathcal{L}_{\text{motion}, i }= -\log \text{prob}_{i,k^*} + \sum_{t=2}^T (\text{pos}_{i,k^*}^t - \text{pos}_{i, *}^t)^2,
\end{equation}
where we encourage the model to have a higher probability for the cloest trajectory $k^*$ and reduce the distance between this trajectory with the ground truth.
The full motion loss is simply:
\begin{equation}
    \mathcal{L}_{\text{motion}}(p, \tau) = \sum_i^N \mathcal{L}_{\text{motion}, i }
\end{equation}
where we sum over all the motion losses for each predicted agent in $p$.

\section{Experiment Details}
\label{supp_exp}

\subsection{Baseline implementation}
\paragraph{TrafficGen~\cite{feng2022trafficgen}.} We use the official implementation\footnote{\url{https://github.com/metadriverse/trafficgen}}.
For a fair comparison, we train its Initialization and Trajectory Generation modules on our dataset for 100 epochs with batch size 64.
We modify $T=50$ in the Trajectory Generation to align with our setting.
We use the default values for all the other hyper-parameters.
During inference, we enforce TrafficGen to generate $N$ vehicles by using the result of the first $N$ autoregressive steps of the Initialization module.

\paragraph{MotionCLIP~\cite{tevet2022motionclip}.}
The core idea of MotionCLIP is to learn a shared space for the interested modality embedding (traffic scenario in our case) and text embedding.
Formally, this model contains a scenario encoder $E$, a text encoder $\hat{E}$, and a scenario decoder $D$. 
For each example of scene-text paired data $(\tau, L, m)$, we encode scenario and text separately with their encoders $\mathbf{z} = E(\tau)$, $\hat{\mathbf{z}} = \hat{E}(L)$.
Then, the decoder takes $\mathbf{z}$ and $m$ and output a scenario $p = D(\mathbf{z}, m)$.
MotionCLIP trains the network with $\mathcal{L}_\text{rec}$ to reconstruct the scenario from the latent code:

\begin{equation}
    \mathcal{L}_\text{rec} = \mathcal{L}{_\text{position}}(p, \tau) + \mathcal{L}_{\text{attr}}(p, \tau) + \mathcal{L}_{\text{motion}}(p, \tau),
\end{equation}
where we use the same set of loss functions as ours (Equation~\ref{equ:loss_ours}).
On the other hand, MotionCLIP aligns the embedding space of the scenario and text with:

\begin{equation}
    \mathcal{L}_\text{align} = 1 - \text{cos}(\mathbf{z}, \hat{\mathbf{z}}),
\end{equation}
which encourages the alignment of scenario embedding $\mathbf{z}$ and text embedding $\hat{\mathbf{z}}$. The final loss function is therefore
\begin{equation}
    \mathcal{L} = \mathcal{L}_\text{rec} + \lambda \mathcal{L}_\text{align},
\end{equation}
where we set $\lambda=100$. 

During inference, given an input text $L$ and a map $m$, we can directly use the text encoder to obtain latent code and decode a scenario from it, formally $\tau = D(\hat{E}(L), m)$.

For the scenario encoder $E$, we use the same scenario encoder as in ~\cite{feng2022trafficgen}, which is a 5-layer \textit{multi-context gating} (MCG) block ~\cite{varadarajan2021multipath} to encode the scene input $\tau$ and outputs $\mathbf{z} \in \mathbb{R}^{1024}$ with the context vector output $c$ of the final MCG block. 
For text encoder $\hat{E}$, we use the sentence embedding of the fixed GPT-2 model.
For the scenario decoder $D$, we modify our \GENERATOR to take in latent representation $\mathbf{z}$ with a dimension of 1024 instead of our own structured representation.
Because $D$ does not receive the number of agents as input, we modify \GENERATOR to produce the $N=32$ agents for every input and additionally add an MLP decoder to predict the objectiveness score of each output agent. 
Here objectiveness score is a binary probability score indicating whether we should put each predicted agent onto the final scenario or not.
During training, for computation of $\mathcal{L}_{\text{rec}}$, we use Hungarian algorithm to pair ground-truth agents with the predicted ones. We then supervise the objectiveness score in a similar way as in DETR.

Note that we need text-scenario paired data to train MotionCLIP. To this end, we use a rule-based method to convert a real dataset $\tau$ to a text $L$. 
This is done by describing different attributes of the scenario with language.
Similar to our Attribute Description dataset, in each text, we enumerate the scenario properties 1) sparsity; 2) position; 3) speed and 4) ego vehicle's motion. 
Here is one example: "the scene is very dense; there exist cars on the front left of ego car; there is no car on the back left of ego car; there is no car on the back right of ego car; there exist cars on the front right of ego car; most cars are moving in fast speed; the ego car stops".

We transform every scenario in our dataset into a text with the format as above. We then train MotionCLIP on our dataset with the same batch size and number of iterations as \MODELNAME.

\subsection{Metric}
We show how to compute MMD in this section.
Specifically, MMD measures the distance between two distributions $q$ and $p$.
\begin{equation}
    \begin{aligned}
        \text{MMD}^2(p,q)= & \mathbb{E}_{x,x' \sim p}[k(x,x')] 
         + \mathbb{E}_{y,y'\sim q}[k(y,y')] \\
         & - 2 \mathbb{E}_{x \sim p, y \sim q}[k(x,y)],
    \end{aligned}
\label{eq:mmd}
\end{equation}

where $k$ is the kernel function (a Gaussian kernel in this work). 
We use Gaussian kernel in this work.
For each pair of real and generated data $(\tau, \hat{\tau})$, we compute the distribution difference between them per attribute.

\subsection{Dataset}
\paragraph{Crash Report.}
We use 38 cases from the CIREN dataset~\cite{national2016crash} from the NHTSA crash report search engine. 
Each case contains a long text description of the scenario as well as a PDF diagram showing the scenario.
Because the texts are very long and require a long time for humans to comprehend, in our human study, along with each text input, we will also show the diagram of the scenario as a reference.
We show example crash reports in Section~\ref{example:input_output}.
We also refer the reader to the NHTSA website~\footnote{\url{https://crashviewer.nhtsa.dot.gov/CIREN/Details?Study=CIREN&CaseId=11}} to view some examples of the crash report.

\paragraph{Attribute Description.} We create text descriptions that highlight various attributes of a traffic scenario. Specifically, we use the following attributes and values:

\begin{enumerate}
    \item Sparsity: "the scenario is \{nearly empty/sparse/with medium density/very dense\}".
    \item Position: "there are only vehicles on the \{left/right/front/back\} side(s) of the center car" or "there are vehicles on different sides of the center car".
    \item Speed: "most cars are moving in \{slow/medium/fast\} speed" or "most cars are stopping".
    \item Ego-vehicle motion: "the center car \{stops/moves straight/turns left/turns right\}".
\end{enumerate}
\begin{figure*}[h]
\centering
% \vspace{-5em}
\includegraphics[width=1.0\linewidth]{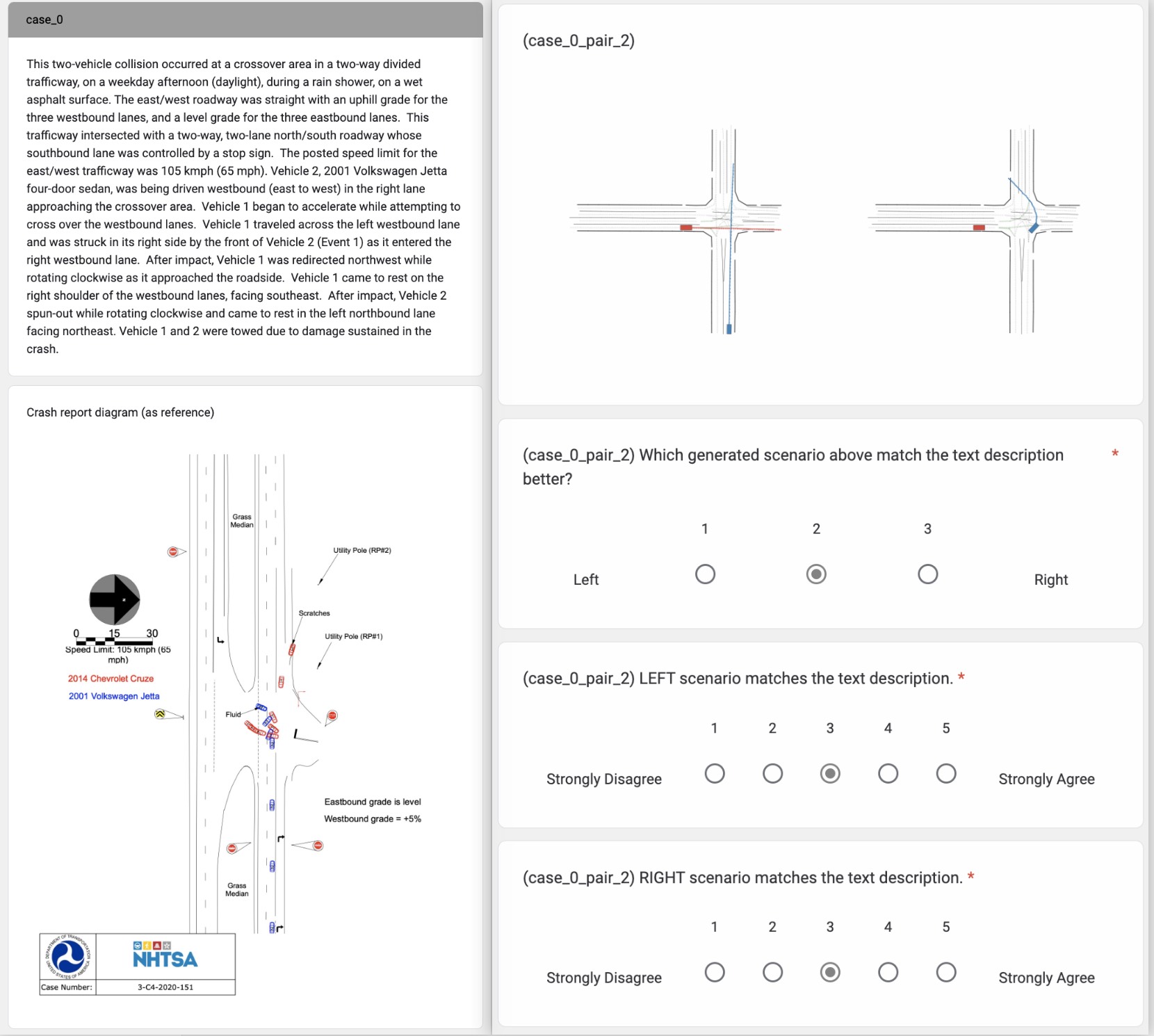}
% \vspace{-3em}
\caption{Human study user interface.}
% \vspace{-1em}
\label{fig:human_study}
\end{figure*}

We create sentences describing each of the single attributes with all the possible values. We also compose more complex sentences by combining 2,3 or 4 attributes together with random values for each of them.
In total, we created 40 cases for human evaluation. Please refer to Section~\ref{example:input_output} for some example input texts from this dataset.

\subsection{Human study}
We conduct the human study to access how well the generated scenario matches the input text.
We showcase the user interface of our human study in Figure~\ref{fig:human_study}.
We compose the output of two models with the same text input in random order and ask the human evaluator to judge which one matches the text description better.
Then, we also ask them to give each output a 1-5 score.
We allow the user to select "unsure" for the first question.

We invite 12 human evaluators for this study, and each of them evaluated all the 78 cases we provided. 
We ensure the human evaluators do not have prior knowledge of how different model works on these two datasets.
On average, the human study takes about 80 minutes for each evaluator.

\subsection{Qualitative result full texts}
In Figure~\ref{fig:qualitative} and Figure~\ref{fig:quali_6}, we show 5 examples of the output of our model on Crash Report data on the first row.
Recall that the texts we show in the figures are the summary from our \INTERPRETER due to space limitations.
We show the full input text for each example in this section.

\setcounter{lstlisting}{0} 
\renewcommand\lstlistingname{Text}
\begin{lstlisting}[breaklines=true,caption={Full texts of examples in Figure~4 .}]
Figure 4 Column 1 (CIREN ID 594):
"This crash occurred during daylight hours on a dry, bituminous divided trafficway (median strip without positive barrier) under clear skies.   There were four east travel lanes (two through lanes, one left turn and one right turn) and four west travel lanes (two through lanes, one left and one right).   The east lanes have a slight right curve and the west lanes curve slightly to the left.  Both east/west travel lanes were level grade at point of impact and divided by a grass median.   The speed limit at this location is 80km/h (50 mph).  The intersecting north/south roadway consisted of one north travel lane and three south travel lanes (one through lanes, one left and one right).  These travel lanes were divided by a raised concrete median on the northern side of the intersection.  This intersection is controlled by overhead traffic signals. A 2017 Dodge Grand Caravan (V1) was traveling east in the left turn lane and a 2006 Nissan Sentra (V2) was traveling west in the left through lane.  As V1 was traveling east it attempted to execute a left turn to travel north when its front bumper impacted the front bumper of V2 (Event 1).   After initial impact, V1 rotated counterclockwise approximately 80 degrees before traveling to its final resting position in the middle of the intersection facing north.  V2 was traveling west in the left through lane and attempting to travel through the intersection when its front bumper impacted the front bumper of V1.   After initial impact V2 rotated clockwise approximately 20 degrees before traveling to its final resting position in the middle of the intersection facing northwest.  V1 and V2 were towed from the scene due to damage sustained in the crash."

Figure 4 Column 2 (CIREN ID 31):
"A 2016 Kia Sedona minivan (V1) was traveling southwest in the right lane of three.  A 2015 Chevrolet Silverado cab chassis pickup (V2) was ahead of V1 in the right lane.  V2 was a working vehicle picking up debris on the highway in a construction zone.  The driver of V2 stopped his vehicle in the travel lane.   The driver of V1 recognized an impending collision and applied the brakes while steering left in the last moment before impact.  V1 slid approximately three meters before the front of V1 struck the back plane of V2 in a rear-end collision with full engagement across the striking planes (Event 1).  Both vehicles came to rest approximately two meters from impact.  V1 was towed due to damage while V2 continued in service."

Figure A2 Row 1 Column 1 (CIREN ID 77):
"A 2017 Chevrolet Malibu LS sedan (V1) was traveling southeast in the right lane cresting a hill. A 1992 Chevrolet C1500 pickup (V2) was traveling northwest in the second lane cresting the same hill. Vehicle 2 crossed left across the center turn lane, an oncoming lane, and then into V1\u2019s oncoming lane of travel. Vehicle 1 and Vehicle 2 collided in a head-on, offset-frontal configuration (Event 1). Vehicle 1 attempted to steer left just before impact, focusing the damage to the middle-right of its front plane. Both vehicles rotated a few degrees clockwise before coming to rest in the roadway, where they were towed from the scene due to damage."

Figure A2 Row 1 Column 2 (CIREN ID 33):
"This two-vehicle collision occurred during the pre-dawn hours (dark, street lights present) of a fall weekday at the intersection of two urban roadways. The crash only involved the eastern leg of the intersection. The westbound lanes of the eastern leg consisted of four westbound lanes that included a right turn lane, two through lanes, and a left turn lane.  The three eastbound lanes of the eastern leg consisted of a merge lane from the intersecting road and two through-lanes. The roadway was straight with a speed limit of 89 kmph (55 mph), and the intersection was controlled by overhead, standard electric, tri-colored traffic signals. At the time of the crash, the weather was clear and the roadway surfaces were dry. As Vehicle 1 approached the intersection, its driver did not notice the vehicles stopped ahead at the traffic light. The traffic signal turned green and Vehicle 2 began to slowly move forward. The frontal plane of Vehicle 1 struck the rear plane of Vehicle 2 (Event 1). Both vehicles came to rest in the left through-lane of the westbound lane facing in a westerly direction. Vehicle 1 was towed from the scene due to damage sustained in the crash. Vehicle 2 was not towed nor disabled. The driver of Vehicle 2 was transported by land to a local trauma center and was treated and released."


Figure A2 Row 1 Column 3 (CIREN ID 56):
"A 2013 Honda CR-V utility vehicle (V1) was traveling west in the right lane approaching an intersection. A 2003 Chevrolet Silverado 1500 pickup (V2) was stopped facing north at a stop sign.   Vehicle 2 proceeded north across the intersection and was struck on the right plane by the front plane of V1 (Event 1). The impact caused both vehicles to travel off the northwest corner of the intersection, where they came to rest. Both vehicles were towed due to damage."

\end{lstlisting}

\section{Additional Results}
\label{supp_result}

\subsection{Controllable self-driving policy evaluation}
\label{supp_result_policy}
We show how \MODELNAME can be utilized to generate interesting scenarios for controllable self-driving policy evaluation. Specifically, we leverage \MODELNAME to generate traffic scenario datasets possessing diverse properties, which we then use to assess self-driving policies under various situations.
For this purpose, we input different text types into \MODELNAME: 1) Crash Report, the real-world crash report data from CIREN; 2) Traffic density specification, a text that describes the scenario as "sparse", "medium dense", or "very dense". For each type of text, we generate 500 traffic scenarios for testing. Additionally, we use 500 real-world scenarios from the Waymo Open dataset.

We import all these scenarios into an interactive driving simulation, MetaDrive~\cite{li2022metadrive}. We evaluate the performance of the IDM~\cite{Treiber_2000} policy and a PPO policy provided in MetaDrive. 
In each scenario, the self-driving policy replaces the ego-vehicle in the scenario and aims to reach the original end-point of the ego vehicle, while all other agents follow the trajectory set out in the original scenario.
We show the success rate and collision rate of both policies in Table~\ref{tab:rl_eval}.
Note that both policies experience significant challenges with the Crash Report scenarios, indicating that these scenarios present complex situations for driving policies. Furthermore, both policies exhibit decreased performance in denser traffic scenarios, which involve more intricate vehicle interactions.
These observations give better insight about the drawbacks of each self-driving policy.
This experiment showcases \MODELNAME as a valuable tool for generating traffic scenarios with varying high-level properties, enabling a more controlled evaluation of self-driving policies.

\begin{table*}[h]
\centering
\begin{tabular}{lccccccc}
\toprule
\multirow{2}{*}{Test Data} & \multicolumn{2}{c}{IDM~\cite{Treiber_2000}} & \multicolumn{2}{c}{PPO (MetaDrive)~\cite{li2022metadrive}} \\
& Success (\%) & Collision (\%) & Success (\%) & Collision (\%) \\\cmidrule{1-5}
Real & 93.60 & 3.80 & 69.32 & 14.67 \\
\midrule
\MODELNAME + \ Crash Report~\cite{national2016crash} & 52.35 & 39.89 & 25.78 & 27.98 \\
\MODELNAME + "Sparse" & 91.03 & 8.21 & 41.03 & 21.06 \\
\MODELNAME + "Medium" & 84.47 & 12.36 & 43.50 & 26.67 \\
\MODELNAME + "Dense" & 68.12 & 19.26 & 38.89 & 32.41 \\
\midrule
\end{tabular}
\caption{Controllable self-driving policy evaluation.}
\label{tab:rl_eval}
\end{table*}

\subsection{Text-conditioned simulation qualitative results}
\label{supp_result_qualitative}
\begin{figure*}
\centering
% \vspace{-4em}
% \includegraphics[width=1.0\linewidth]{figures/result_figures/text_quali.pdf}
\includegraphics[width=1.0\linewidth]{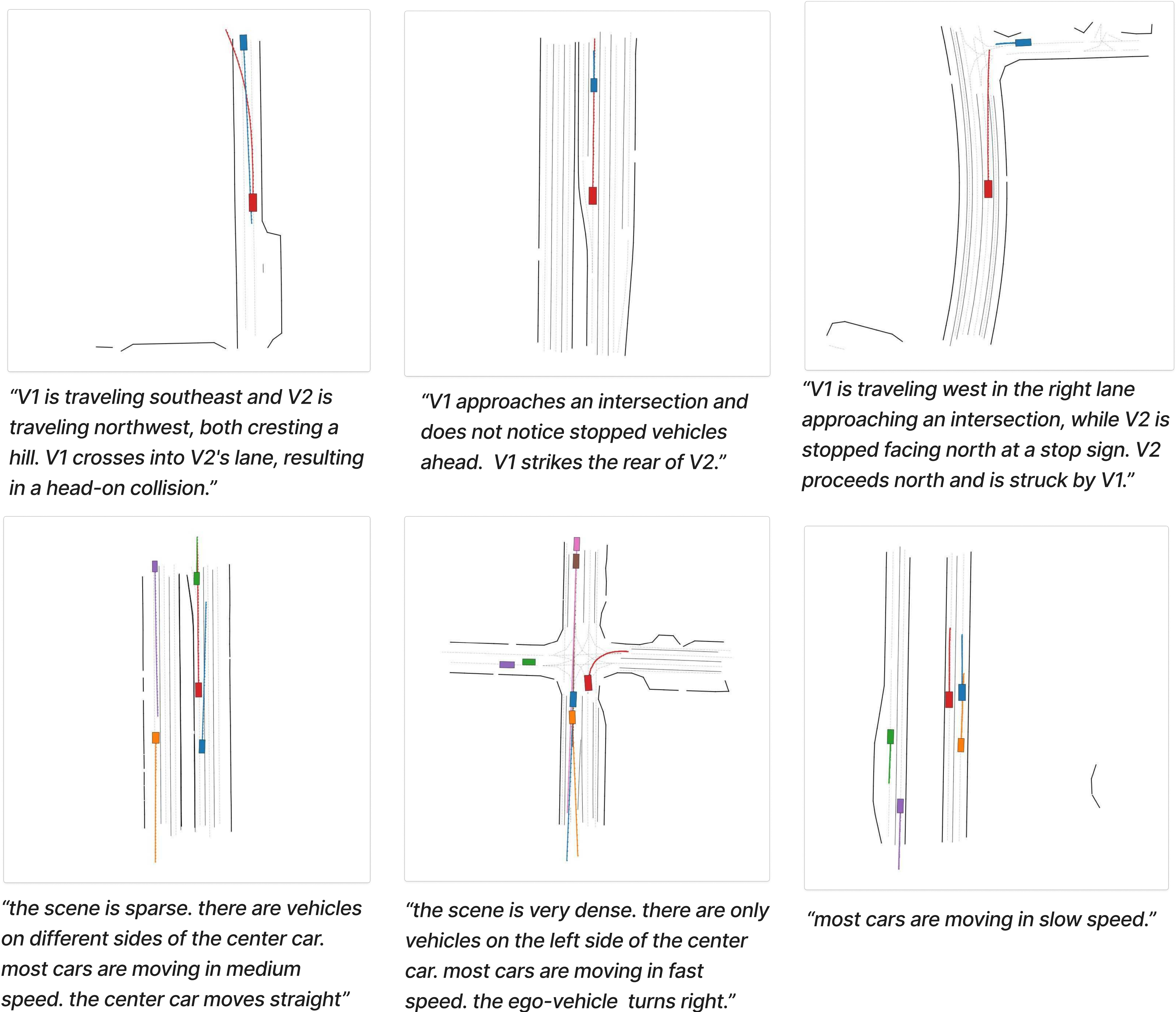}
% \vspace{-4em}
\caption{Qualitative results on text-conditioned generation.}
% \vspace{-2em}
\label{fig:quali_6}
\end{figure*}
We show the more qualitative results of text-conditioned simulation in Figure~\ref{fig:quali_6}.
Here, the upper 3 examples are from the Crash Report dataset, the lower 3 examples are from the Attribute Description dataset.

\subsection{Human study statistics}
\label{supp_result_human_study}
\paragraph{Score distribution.}
\begin{figure}
\centering
\begin{subfigure}{.5\textwidth}
  \centering
  \includegraphics[width=\linewidth]{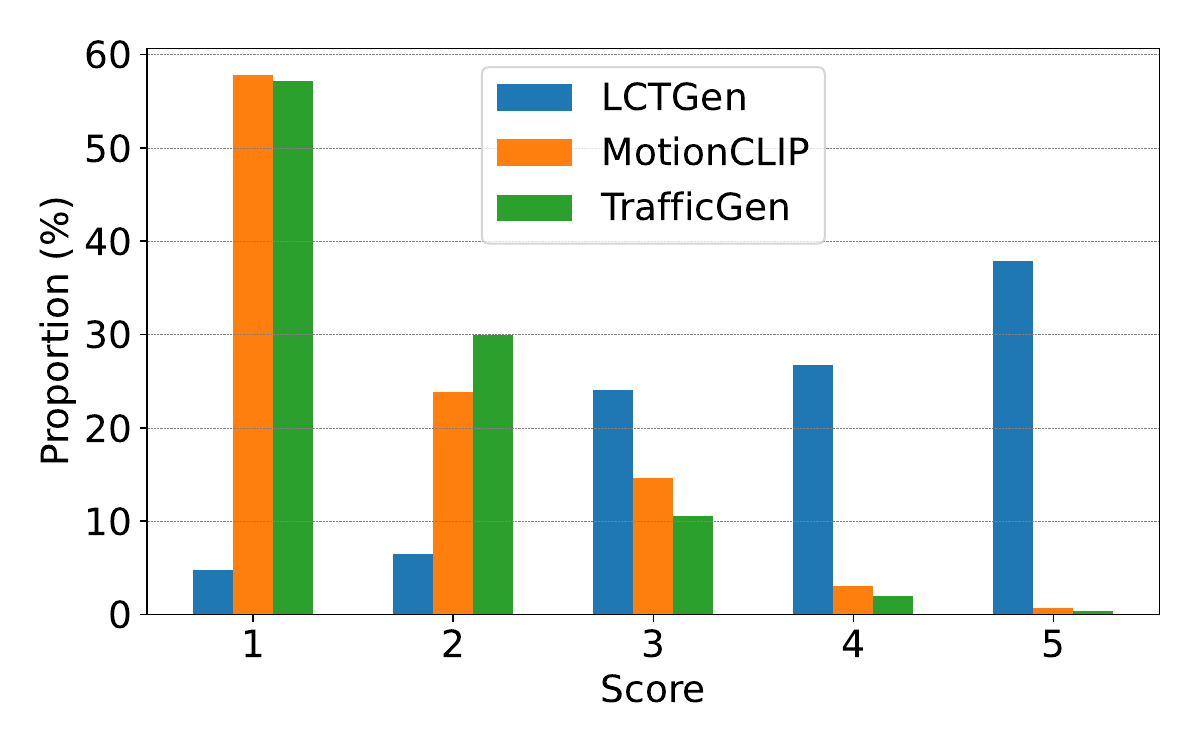}
  \caption{Crash Report}
  \label{fig:sub1}
\end{subfigure}%
\begin{subfigure}{.5\textwidth}
  \centering
  \includegraphics[width=\linewidth]{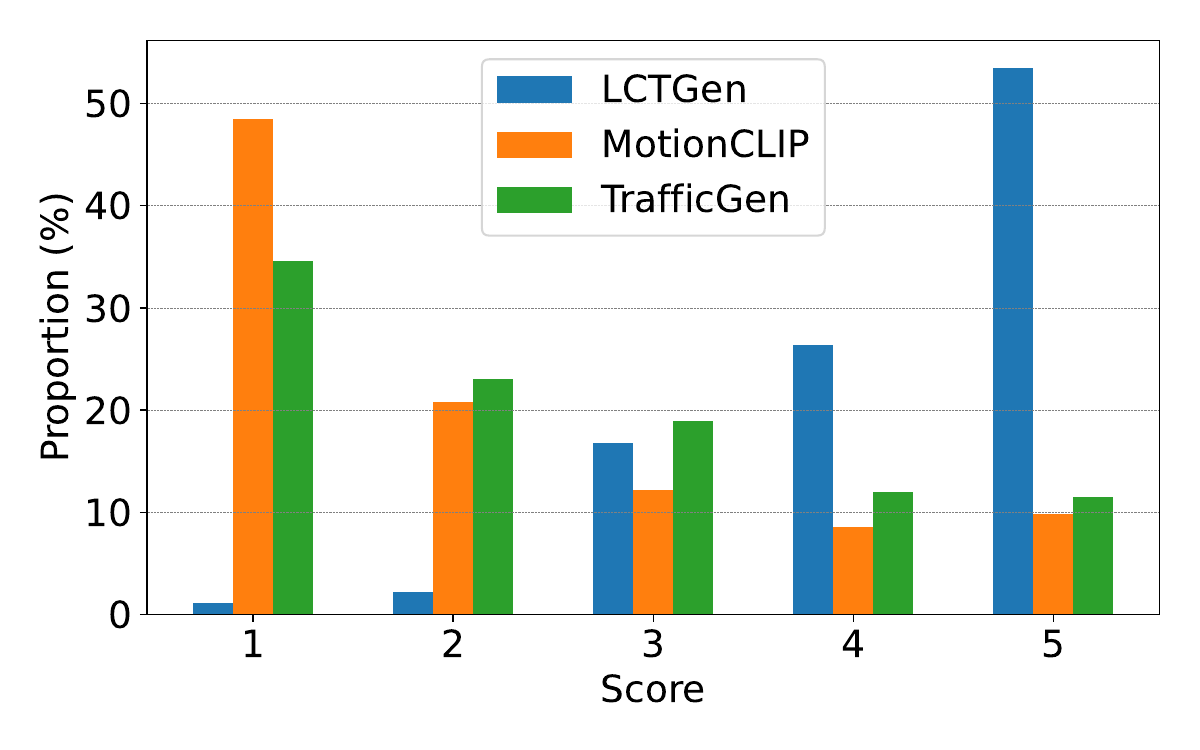}
  \caption{Attribute Description}
  \label{fig:sub2}
\end{subfigure}
\caption{Human study score distribution.}
\label{fig:human_score_dist}
\end{figure}
We show the human evaluation scores of the two datasets in Figure~\ref{fig:human_score_dist}.
We observe that our method is able to reach significantly better scores from human evaluators.
\paragraph{A/B test distribution.}
\begin{figure}
\centering
\begin{subfigure}{1.00\textwidth}
  \centering
  \includegraphics[width=0.32\linewidth]{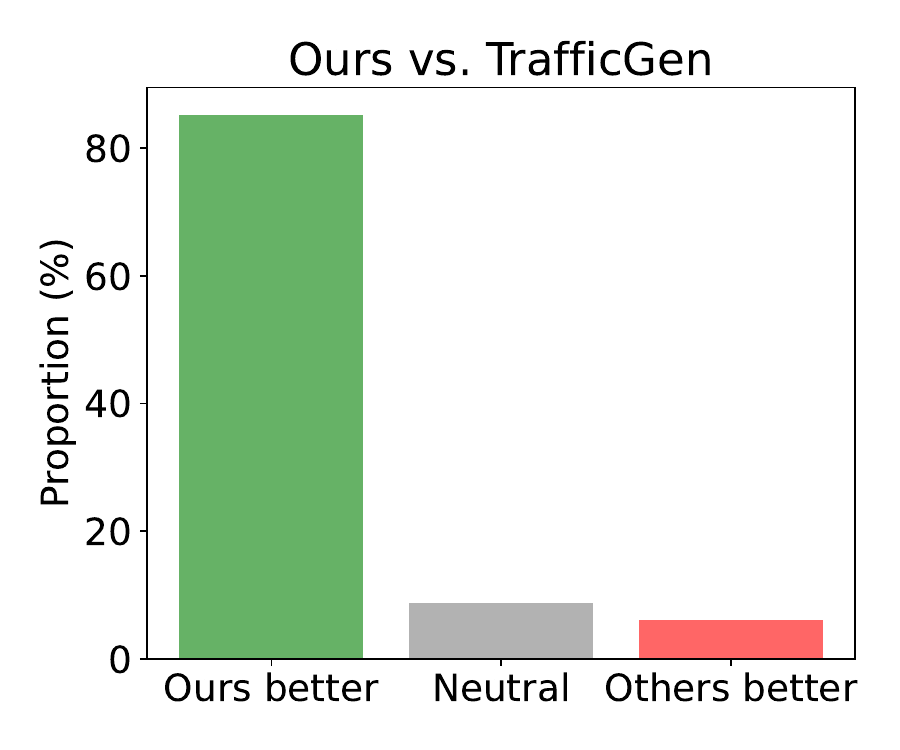}
  \includegraphics[width=0.32\linewidth]{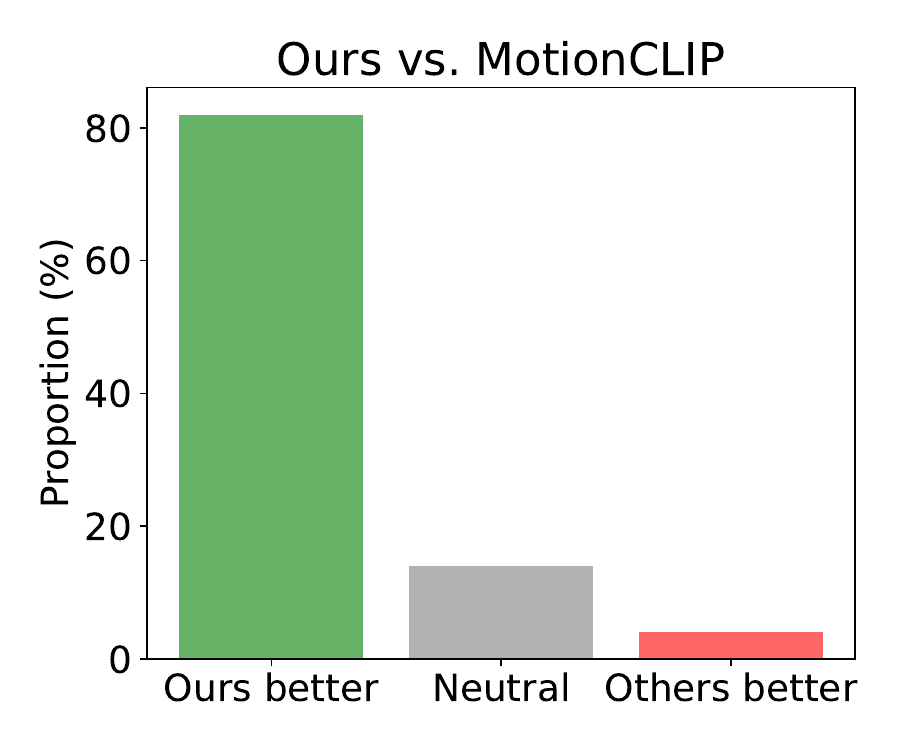}
  \includegraphics[width=0.32\linewidth]{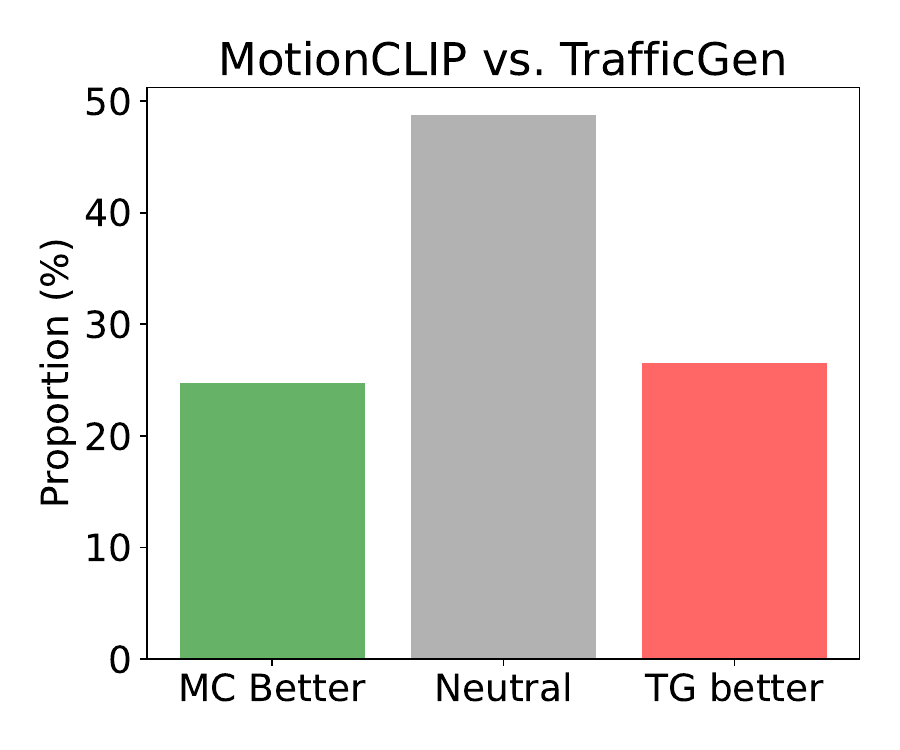}
  \caption{Crash Report}
\end{subfigure}%
\\
\begin{subfigure}{1\textwidth}
  \centering
  \includegraphics[width=0.32\linewidth]{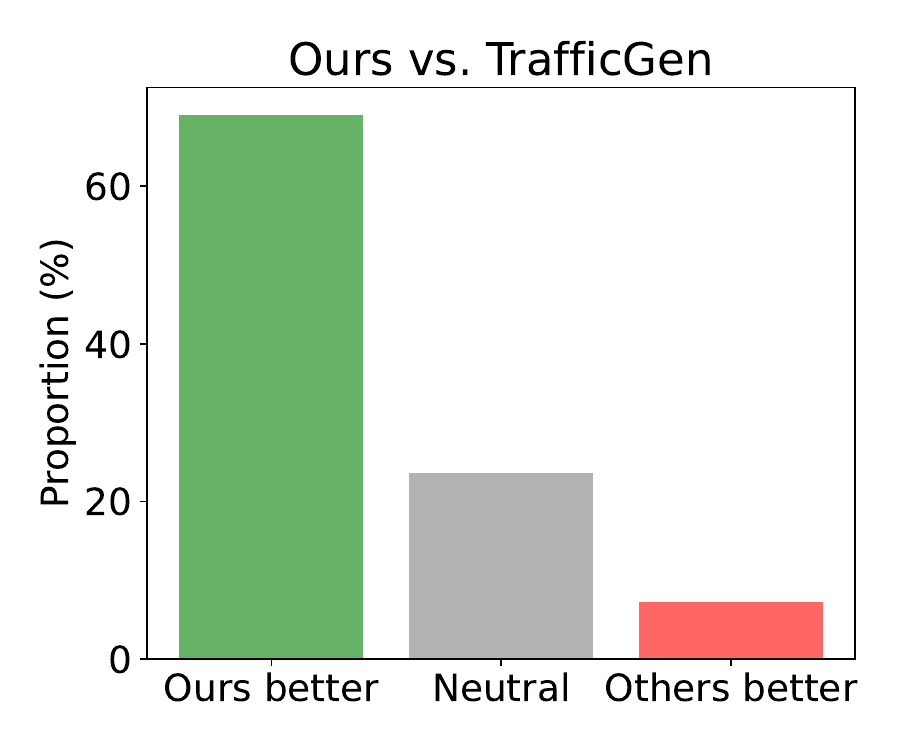}
  \includegraphics[width=0.32\linewidth]{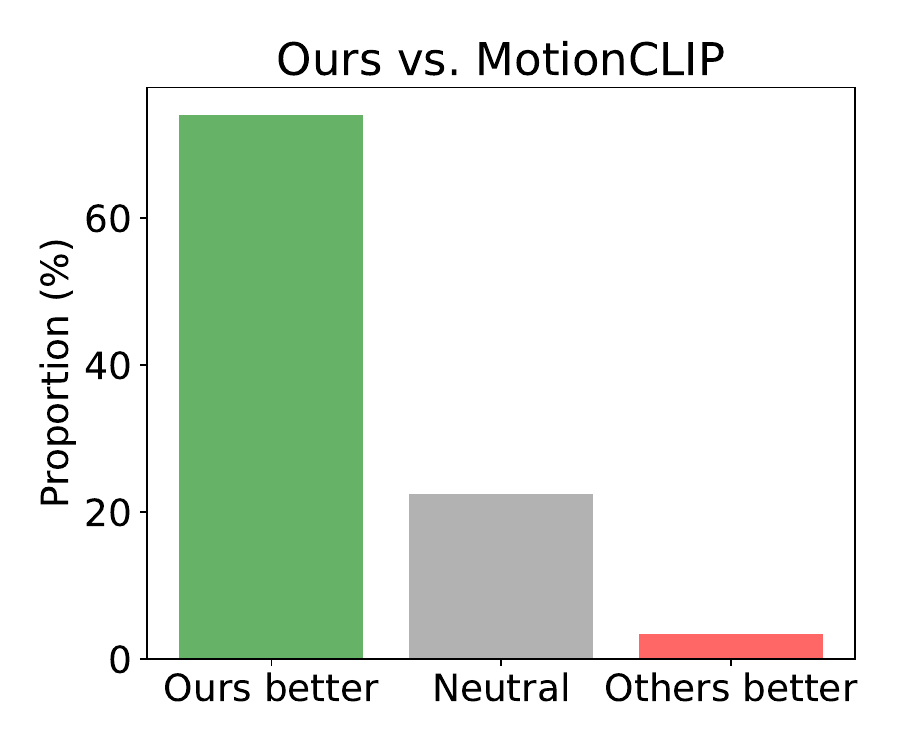}
  \includegraphics[width=0.32\linewidth]{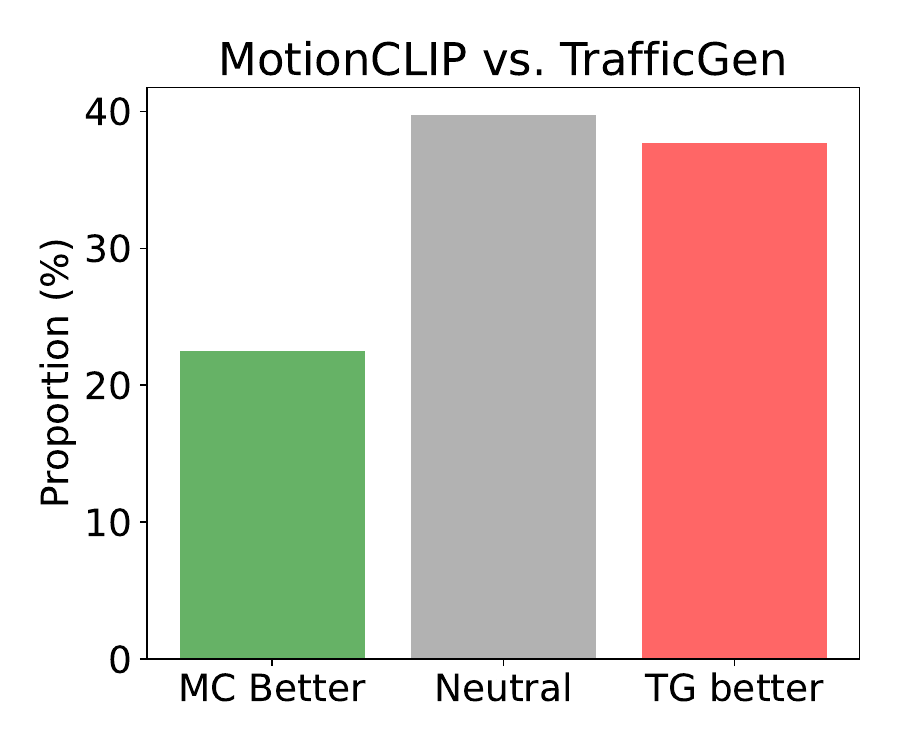}
  \caption{Attribute Description}
\end{subfigure}
\caption{Human study A/B test distribution.}
\label{fig:ab_test}
\end{figure}
We show the distribution of A/B test result for each pair of methods in Figure~\ref{fig:ab_test}.
Note that our method is chosen significantly more frequenly as the better model compared with other models.
We also observe that TrafficGen is slightly better than MotionCLIP in Attribute Description dataset, while the two models achieve similar results in Crash Report.

\paragraph{Human score variance.}
\begin{table*}[t]
\centering
\begin{tabular}{lccccccc}
\toprule
\multirow{2}{*}{Method} & \multicolumn{2}{c}{Crash Report} & \multicolumn{2}{c}{Attribute Description} \\
& Avg. Score & Human Std. & Avg. Score & Human Std.
\\\cmidrule{1-5}
TrafficGen~\cite{feng2022trafficgen} & 1.58 & 0.64 & 2.43 & 0.72 \\
MotionCLIP~\cite{tevet2022motionclip} & 1.65 & 0.67 & 2.10 & 0.64 \\
\midrule
\MODELNAME & 3.86 & 0.87 & 4.29 & 0.65 \\
\midrule
\end{tabular}
\caption{Human study average score and variance.}
\label{tab:human_std}
\end{table*}

We show the variance of quality score across all human evaluators in Table~\ref{tab:human_std}.
Specifically, for each case, we compute the standard deviation across all the human evaluators for this case. 
Then, we average all the standard deviation values across all the cases and show in the table as "Human Std.".
This value measures the variance of score due to human evaluators' subjective judgement differences.
According to the average score and human variance shown in the table, we conclude that our model outperforms the compared methods with high confidence levels.

\subsection{\INTERPRETER input-output examples}
\label{example:input_output}

Here we show the full-text input and output of \INTERPRETER for four examples in Figure~4.
Specifically, we show two examples from Crash Report and two examples from Attribute Descriptions.

\vspace{1.0em}

\begin{lstlisting}[breaklines=true,caption={Input-output examples of \INTERPRETER.}]
Figure 4 Column 1 (CIREN ID 594):
Input:
"This crash occurred during daylight hours on a dry, bituminous divided trafficway (median strip without positive barrier) under clear skies.   There were four east travel lanes (two through lanes, one left turn and one right turn) and four west travel lanes (two through lanes, one left and one right).   The east lanes have a slight right curve and the west lanes curve slightly to the left.  Both east/west travel lanes were level grade at point of impact and divided by a grass median.   The speed limit at this location is 80km/h (50 mph).  The intersecting north/south roadway consisted of one north travel lane and three south travel lanes (one through lanes, one left and one right).  These travel lanes were divided by a raised concrete median on the northern side of the intersection.  This intersection is controlled by overhead traffic signals. A 2017 Dodge Grand Caravan (V1) was traveling east in the left turn lane and a 2006 Nissan Sentra (V2) was traveling west in the left through lane.  As V1 was traveling east it attempted to execute a left turn to travel north when its front bumper impacted the front bumper of V2 (Event 1).   After initial impact, V1 rotated counterclockwise approximately 80 degrees before traveling to its final resting position in the middle of the intersection facing north.  V2 was traveling west in the left through lane and attempting to travel through the intersection when its front bumper impacted the front bumper of V1.   After initial impact V2 rotated clockwise approximately 20 degrees before traveling to its final resting position in the middle of the intersection facing northwest.  V1 and V2 were towed from the scene due to damage sustained in the crash."

Output:
Summary: V1 is traveling east in the left turn lane and attempts to turn left when it collides with V2 traveling west in the left through lane. There are 2 vehicles in this scenario. This happens at the intersection of an eight-lane divided trafficway and a four-lane north/south roadway, controlled by traffic signals.
Explanation: 
- V1 (ego vehicle) is traveling east in the left turn lane and attempts to turn left. We cannot find V1's speed in the query. Because V1 tries to turn left, its initial speed should be set low. We set V1's speed as 5 m/s, which has the index of 2. V1 turns left, so its actions are all 1 (turn left).
- V2 is traveling west in the left through lane. As V1 is turning left, 5 seconds before the crash, V2 is coming from the opposite direction (westbound), crossing the path of V1. In the coordinates of V1 (which is facing east initially), V2 comes from the front and is on the left side. Hence, V2's position is \"front left\" (0). As V1 is facing east and V2 facing west, V2 is moving in the parallel opposite direction with V1. Therefore its direction is 1 (parallel_opposite). We cannot find V2's speed in the query. Because V2 is traveling west and hit by V1 5 seconds later, we assume V2's init speed is 8 m/s (index 3). Given this speed, V2's distance to V1 is 8m/s * 5s = 40m (index 8). V2 keeps going straight, so its actions are all 4 (keep speed).
- Map: This happens at the intersection of an eight-lane divided trafficway (4 lanes for eastbound and westbound traffic) and a four-lane north/south roadway (1 lane for northbound and 3 lanes for southbound traffic). The ego vehicle is in the left turn-only lane with lane id 4. There is an intersection with a four-lane north/south road ahead. V1 is about to cross the interaction, therefore the distance to the intersection is with 15 meter index 3.
Actor Vector:
- 'V1': [-1, 0, 0, 2, 1, 1, 1, 1]
- 'V2': [0, 8, 1, 3, 4, 4, 4, 4]
Map Vector:
- 'Map': [4, 4, 1, 3, 3, 4]

Figure A2 Row 1 Column 2 (CIREN ID 33):
Input:
"This two-vehicle collision occurred during the pre-dawn hours (dark, street lights present) of a fall weekday at the intersection of two urban roadways. The crash only involved the eastern leg of the intersection. The westbound lanes of the eastern leg consisted of four westbound lanes that included a right turn lane, two through lanes, and a left turn lane.  The three eastbound lanes of the eastern leg consisted of a merge lane from the intersecting road and two through-lanes. The roadway was straight with a speed limit of 89 kmph (55 mph), and the intersection was controlled by overhead, standard electric, tri-colored traffic signals. At the time of the crash, the weather was clear and the roadway surfaces were dry. As Vehicle 1 approached the intersection, its driver did not notice the vehicles stopped ahead at the traffic light. The traffic signal turned green and Vehicle 2 began to slowly move forward. The frontal plane of Vehicle 1 struck the rear plane of Vehicle 2 (Event 1). Both vehicles came to rest in the left through-lane of the westbound lane facing in a westerly direction. Vehicle 1 was towed from the scene due to damage sustained in the crash. Vehicle 2 was not towed nor disabled. The driver of Vehicle 2 was transported by land to a local trauma center and was treated and released."

Output:
Summary: V1 approaches an intersection and does not notice stopped vehicles ahead. The traffic light turns green, and V2 starts moving forward slowly. V1 strikes the rear of V2. There are 2 vehicles in this scenario. This happens at the intersection of two urban roadways with multiple lanes and traffic signals.
Explanation: 
- V1 (ego vehicle) is approaching the intersection and does not notice the stopped vehicles ahead. We cannot find V1's speed in the query. Because V1 is moving and does not stop, we assume V1's initial speed is 10 m/s (index 4). V1 keeps going straight, so its actions are all 4 (keep speed).
- V2 is stopped at the traffic light and starts moving forward slowly when the light turns green. As V1 is moving straight and hits V2 from behind, 5 seconds before the crash, V2 is in front of V1. Hence, V2's position is \"front right\" (3). As V1 and V2 are both moving westbound, V2 is moving in the parallel same direction with V1. Therefore its direction is 0 (parallel_same). V2's speed is slow, so we set it to 2.5 m/s (index 1). Given this speed, V2's distance to V1 is 10m/s * 5s - 2.5m/s * 5s = 37.5m (index 7). V2 keeps going straight, so its actions are all 4 (keep speed).
- Map: This happens at the intersection of two urban roadways with multiple lanes and traffic signals. The ego vehicle is in the left through-lane of the westbound lane with lane id 3. There are 4 westbound lanes and 3 eastbound lanes. The distance of the ego vehicle to the intersection is set to be 10m/s * 5s = 50m (index 10).
Actor Vector:
- 'V1': [-1, 0, 0, 4, 4, 4, 4, 4]
- 'V2': [3, 7, 0, 1, 4, 4, 4, 4]
Map Vector:
- 'Map': [4, 3, 0, 0, 10, 3]


Figure A2 Row 2 Column 1:
Input:
"the scene is with sparse density. there are vehicles on different sides of the center car. most cars are moving in medium speed. the center car moves straight"

Output:
Summary: The ego vehicle (V1) moves straight at a medium speed in a sparse traffic scenario. There are vehicles on different sides of the ego vehicle, corresponding to the sparse density. We generate 4 vehicles in the scene on a six-lane two-way road. The ego car is placed on the third lane from the right.
Explanation:
- V1 (ego vehicle) is in the center and moves straight. Because V1 moves with a medium speed, its speed is set to 4 (medium speed) and action indices are 4 (keep speed).
- V2 is in the closest distance range (index 0). It is moving in the same direction as the ego vehicle (index 0) and is located in the back right (index 2) of the ego vehicle. It is moving with a medium speed (index 4), and its actions are all to keep speed (index 4).
- V3 is in a further distance range (index 1). It is moving in the opposite direction of the ego vehicle (index 1), therefore it is on the left side of the ego vehicle, locating in the back left (index 1) of the ego vehicle. It is moving with a medium speed (index 4), and its actions are all to keep speed (index 4).
- V4 is in the farthest distance range (index 2). It is moving in the same direction as the ego vehicle (index 0) and is located in the front right (index 3) of the ego vehicle. It is moving with a medium speed (index 4), and its actions are all to keep speed (index 4).
- V5 is in the farthest distance range (index 2). It is moving in the opposite direction of the ego vehicle (index 1), therefore it is on the left side of the ego vehicle, locating in the front left (index 0) of the ego vehicle. It is moving with a medium speed (index 4), and its actions are all to keep speed (index 4).
- Map: Given the sparse density and the presence of vehicles on different sides of the ego vehicle, we assume this happens on a 6-lane two-way road. The ego vehicle is in the third lane from the right, with lane id 3. There is no intersection nearby, indicated by -1.
Actor Vector:
- 'V1': [-1, 0, 0, 4, 4, 4, 4, 4]
- 'V2': [2, 0, 0, 4, 4, 4, 4, 4]
- 'V3': [1, 1, 1, 4, 4, 4, 4, 4]
- 'V4': [3, 2, 0, 4, 4, 4, 4, 4]
- 'V5': [0, 2, 1, 4, 4, 4, 4, 4]
Map Vector:
- 'Map': [3, 3, 0, 0, -1, 3]

Figure 4 Column 3:
Input:
"the scene is with sparse density. there are only vehicles on the back side of the center car. most cars are moving in fast speed. the center car turns left"

Output:

Summary: The ego vehicle (V1) turns left at a fast speed in a sparse traffic scenario. All other vehicles are located behind the ego vehicle, which corresponds to the scenario's sparse density. We generate 3 vehicles in the scene on a six-lane two-way road with an intersection in front. The ego car is placed on the left-turn lane.
Explanation:
- V1 (ego vehicle) is in the center and turns left. Because V1 moves with a fast speed, its speed is set to 6 (fast speed) and action indices are 1 (turn left).
- V2-V3 are in the closest distance range (index 0). They are moving in the same direction as the ego vehicle (index 0). Because the ego car is on the left-turn lane, they are both located in the back right (index 2) of the ego vehicle. They are both moving with a fast speed (index 6 or 7), and their actions are all to keep speed (index 4).
- V4 is in a further distance range (index 1). It is moving in the opposite direction of the ego vehicle (index 1), therefore it is on the left side of the ego vehicle, locating in the back left (index 1) of the ego vehicle. It is moving with a fast speed (index 6), in the opposite direction as the ego vehicle (index 1). Its actions are all to keep speed (index 4).
- Map: Given the sparse density and the presence of vehicles behind the ego vehicle, we assume this happens on a 6-lane two-way road. The ego vehicle is in the left-turn lane, with lane id 3. There is an intersection 10 meters ahead (index 2) as the ego vehicle is turning left.
Actor Vector:
- 'V1': [-1, 0, 0, 6, 1, 1, 1, 1]
- 'V2': [2, 0, 0, 6, 4, 4, 4, 4]
- 'V3': [2, 0, 0, 7, 4, 4, 4, 4]
- 'V4': [1, 1, 1, 6, 4, 4, 4, 4]
Map Vector:
- 'Map': [3, 3, 2, 2, 2, 3]
\end{lstlisting}

\subsection{Attribute Description Result Split}
\label{supp_result_ablation}
\begin{table*}[h]
\centering
\begin{tabular}{lccccccc}
\toprule
Method & Density & Position & Speed & Ego-car Motion\\\cmidrule{1-5}
TrafficGen~\cite{feng2022trafficgen} & 2.75 & 2.03 & 2.34 & 2.27 \\
MotionCLIP~\cite{tevet2022motionclip} & 1.89 & 2.24 & 1.91 & 1.78 \\
\midrule
\MODELNAME & \textbf{4.24} & \textbf{4.28} & \textbf{4.38} & \textbf{4.40} \\
\midrule
\end{tabular}
\caption{Human study result split analysis on Attribute Description scores.}
\label{tab:human_study_split}
\end{table*}

We generate the Attribute Description dataset with different attributes.
In this section, we split the matching score result for the full dataset into different attributes.
We show the result in Table~\ref{tab:human_study_split}.
We observe our method has nearly identical performance over all the attributes.
TrafficGen the best results with Density, while MotionCLIP performs the best with Position.

\subsection{Full Ablation Study} 
\label{supp_result_inst}
\begin{table*}[h]
\centering
\begin{tabular}{lccccccc}
\toprule
\multirow{2}{*}{Method} & \multicolumn{4}{c}{Initialization} & \multicolumn{3}{c}{Motion} \\
& Pos & Heading & Speed & Size & mADE &  mFDE & SCR  \\\cmidrule{1-8}
w/o Quad. & 0.092 & 0.122 & 0.076 & 0.124 & 2.400 & 4.927 & 8.087 \\
w/o Dist. & 0.071 & 0.124 & 0.073 & 0.121 & 1.433 & 3.041 & 6.362 \\
w/o Ori. & 0.067 & 0.132 & 0.082 & 0.122 & 1.630 & 3.446 & 7.300 \\
w/o Speed & 0.063 & 0.120 & 0.104 & 0.122 & 2.611 & 5.188 & 7.150 \\
w/o Action & 0.067 & 0.128 & 0.173 & 0.128 & 2.188 & 5.099 & 7.146 \\
w/o $x_i$ & 0.067 & 0.133 & 0.076 & 0.124 & 1.864 & 3.908 & \textbf{5.929} \\
w/o GMM & 0.064 & 0.128 & 0.078 & 0.178 & 1.606 & 3.452 & 8.216 \\
\midrule
\MODELNAME & \textbf{0.062} & \textbf{0.115} & \textbf{0.072} & \textbf{0.120} & \textbf{1.329} & \textbf{2.838} & 6.700 \\
\midrule
\end{tabular}
\caption{Ablation study of \MODELNAME
}
\label{tab:abl_full}
\end{table*}

In our main paper, we split the ablation study into two different groups. Here we show the full results of all the ablated methods in Table~\ref{tab:abl_full}.
We additionally show the effect of 1) using the learnable query $x_i$ and 2) using the GMM prediction for attributes.

\subsection{Instructional traffic scenario editing}
\begin{figure*}[h]
\centering
\includegraphics[width=1.0\linewidth]{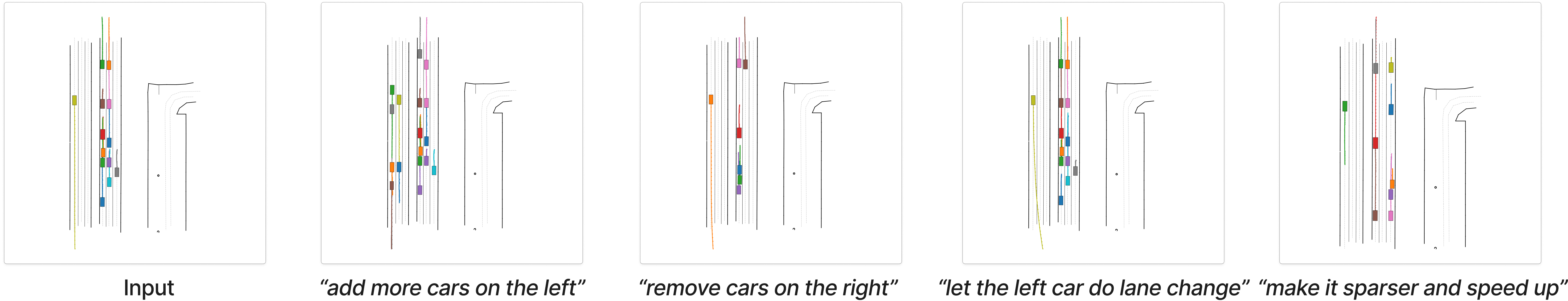}
\caption{Instructional editing on a real-world scenario}
\label{fig:edit_2}
\end{figure*}
We show another example of instructional traffic scenario editing in Figure~\ref{fig:edit_2}.
Different from the compound editing in Figure~\ref{fig:edit} in the main paper, here every example is edited from the input scenario.

% \bibliography{egbib}  % .bib

\end{document}